# An Approach to Temporal Planning and Scheduling in Domains with Predictable Exogenous Events

**Alfonso Gerevini**                                            GEREVINI@ING.UNIBS.IT
**Alessandro Saetti**                                            SAETTI@ING.UNIBS.IT
**Ivan Serina**                                            SERINA@ING.UNIBS.IT
*Dipartimento di Elettronica per l'Automazione*
*Università degli Studi di Brescia*
*Via Branze 38, I-25123 Brescia, Italy*

## Abstract

The treatment of exogenous events in planning is practically important in many real-world domains where the preconditions of certain plan actions are affected by such events. In this paper we focus on planning in temporal domains with exogenous events that happen at known times, imposing the constraint that certain actions in the plan must be executed during some predefined time windows. When actions have durations, handling such temporal constraints adds an extra difficulty to planning. We propose an approach to planning in these domains which integrates constraint-based temporal reasoning into a graph-based planning framework using local search. Our techniques are implemented in a planner that took part in the 4th International Planning Competition (IPC-4). A statistical analysis of the results of IPC-4 demonstrates the effectiveness of our approach in terms of both CPU-time and plan quality. Additional experiments show the good performance of the temporal reasoning techniques integrated into our planner.

## 1. Introduction

In many real-world planning domains, the execution of certain actions can only occur during some predefined time windows where one or more necessary conditions hold. For instance, a car can be refueled at a gas station only when the gas station is open, or a space telescope can take a picture of a certain planet region only when this region is observable. The truth of these conditions is determined by some exogenous events that happen at known times, and that cannot be influenced by the actions available to the planning agent (e.g., the closing of the gas station or the planet movement).

Several frameworks supporting action durations and time windows have been proposed (e.g., Vere, 1983; Muscettola, 1994; Laborie & Ghallab, 1995; Schwartz & Pollack, 2004; Kavuluri & U, 2004; Sanchez, Tang, & Mali, 2004). However, most of them are domain-dependent systems or are not fast enough on large-scale problems. In this paper, we propose a new approach to planning with these temporal features, integrating constraint-based temporal reasoning into a graph-based planning framework.

The last two versions of the domain definition language of the International planning competition (IPC) support action durations and predictable (deterministic) exogenous events (Fox & Long, 2003; Edelkamp & Hoffmann, 2004). In PDDL2.1, predictable exogenous events can be implicitly represented (Fox, Long, & Halsey, 2004), while in PDDL2.2 they can be explicitly represented through *timed initial literals*, one of the two new PDDL features on which the 2004 competition (IPC-4) focused. Timed initial literals are specified in the description of the initial state of the planning problem through assertions of the form "(at t L)", where t is a real number, and L is a ground literal whose predicate does not appear in the effects of any domain action. The obvious meaning of (at t L) is that L is true from time t. A set of these assertions involving the same ground predicate defines a sequence of disjoint time windows over which the timed predicate holds. An example in the well-known "ZenoTravel" domain (Penberthy, 1993; Long & Fox, 2003a) is

```
(at 8 (open-fuelstation city1))
(at 12 (not (open-fuelstation city1)))
(at 15 (open-fuelstation city1))
(at 20 (not (open-fuelstation city1))).
```

These assertions define two time windows over which (open-fuelstation city1) is true, i.e., from 8 to 12 (excluded) and from 15 to 20 (excluded). A timed initial literal is relevant to the planning process when it is a precondition of a domain action, which we call a *timed precondition* of the action. Each timed precondition of an action can be seen as a temporal scheduling constraint for the action, defining the feasible time window(s) when the action can be executed. When actions in a plan have durations and timed preconditions, computing a valid plan requires planning and reasoning about time to be integrated, in order to check whether the execution of the planned actions can satisfy their scheduling constraints. If an action in the plan cannot be scheduled, then the plan is not valid and it must be revised.

The main contributions of this work are: (i) a new representation of temporal plans with action durations and timed preconditions, called *Temporally-Disjunctive Action Graph*, (TDA-graph) integrating disjunctive constraint-based temporal reasoning into a recent graph-based approach to planning; (ii) a polynomial method for solving the disjunctive temporal reasoning problems that arise in this context; (iii) some new local search techniques to guide the planning process using our representation; and (iv) an experimental analysis evaluating the performance of our methods implemented in a planner called LPG-td, which took part in IPC-4 showing very good performance in many benchmark problems.

The "*td*" extension in the name of our planner is an abbreviation of "*t*imed initial literals and *d*erived predicates", the two main new features of PDDL2.2.[1] In LPG-td, the techniques for handling timed initial literals are quite different from the techniques for handling derived predicates. The first ones concern representing temporal plans with predictable exogenous events and fast temporal reasoning for action scheduling during planning; the second ones concern incorporating a rule-based inference system for efficient reasoning about derived predicates during planning. Both timed initial literals and derived predicates require to change the heuristics guiding the search of the planner, but in a radically different way. In this paper, we focus on timed initial literals, which are by themselves a significant and useful extension to PDDL2.1. Moreover, an analysis of the results of IPC-4 shows that LPG-td was top performer in the benchmark problems involving this feature. The treatment of derived predicates in LPG-td is presented in another recent paper (Gerevini et al., 2005b).

---

1. Derived predicates allow us to express in a concise and natural way some indirect action effects. Informally, they are predicates which do not appear in the effect of any action, and their truth is determined by some domain rules specified as part of the domain description.

The paper is organized as follows. In Section 2, after some necessary background, we introduce the TDA-graph representation and a method for solving the disjunctive temporal reasoning problems that arise in our context. In Section 3, we describe some new local search heuristics for planning in the space of TDA-graphs. In Section 4, we present the experimental analysis illustrating the efficiency of our approach. In Section 5, we discuss some related work. Finally, in Section 6 we give the conclusions.

## 2. Temporally Disjunctive Action Graph

Like in partial-order causal-link planning, (e.g., Penberthy & Weld, 1992; McAllester & Rosenblitt, 1991; Nguyen & Kambhampati, 2001), in our framework we search in a space of partial plans. Each search state is a partial temporal plan that we represent by a Temporally-Disjunctive Action Graph (TDA-graph). A TDA-graph is an extension of the linear action graph representation (Gerevini, Saetti, & Serina, 2003) which integrates disjunctive temporal constraints for handling timed initial literals. A linear action graph is a variant of the well-known planning graph (Blum & Furst, 1997). In this section, after some necessary background on linear action graphs and disjunctive temporal constraints, we introduce TDA-graphs, and we propose some techniques for temporal reasoning in the context of this representation that will be used in the next section.

### 2.1 Background: Linear Action Graph and Disjunctive Temporal Constraints

A linear action graph (LA-graph) $\mathcal{A}$ for a planning problem $\Pi$ is a directed acyclic leveled graph alternating a *fact level*, and an *action level*. Fact levels contain *fact nodes*, each of which is labeled by a ground predicate of $\Pi$. Each fact node $f$ at a level $l$ is associated with a *no-op* action node at level $l$ representing a dummy action having the predicate of $f$ as its only precondition and effect. Each action level contains one action node labeled by the name of a domain action that it represents, and the no-op nodes corresponding to that level.

An action node labeled $a$ at a level $l$ is connected by incoming edges from the fact nodes at level $l$ representing the preconditions of $a$ (*precondition nodes*), and by outgoing edges to the fact nodes at level $l + 1$ representing the effects of $a$ (*effect nodes*). The initial level contains the special action node $a_{start}$, and the last level the special action node $a_{end}$. The effect nodes of $a_{start}$ represent the positive facts of the initial state of $\Pi$, and the precondition nodes of $a_{end}$ the goals of $\Pi$.

A pair of action nodes (possibly no-op nodes) can be constrained by a *persistent mutex relation* (Fox & Long, 2003), i.e., a mutually exclusive relation holding at every level of the graph, imposing that the involved actions can never occur in parallel in a valid plan. Such relations can be efficiently precomputed using an algorithm that we proposed in a previous work (Gerevini et al., 2003).

An LA-graph $\mathcal{A}$ also contains a set of *ordering constraints* between actions in the (partial) plan represented by the graph. These constraints are (i) constraints imposed during search to deal with mutually exclusive actions: if an action $a$ at level $l$ of $\mathcal{A}$ is mutex with an action node $b$ at a level after $l$, then $a$ is constrained to finish before the start of $b$; (ii) constraints between actions implied by the causal structure of the plan: if an action $a$ is used to achieve a precondition of an action $b$, then $a$ is constrained to finish before the start of $b$.

The effects of an action node can be automatically propagated to the next levels of the graph through the corresponding no-ops, until there is an interfering (mutex) action "blocking" the propagation, or the last level of the graph has been reached (Gerevini et al., 2003). In the rest of the paper, we assume that the LA-graph incorporates this propagation.

A *Disjunctive Temporal Problem* (DTP) (Stergiou & Koubarakis, 2000; Tsamardinos & Pollack, 2003) is a pair $\langle \mathcal{P}, \mathcal{C} \rangle$, where $\mathcal{P}$ is a set of time point variables, $\mathcal{C}$ is a set of disjunctive constraints $c_1 \vee \cdots \vee c_n$, $c_i$ is of form $y_i - x_i \leq k_i$, $x_i$ and $y_i$ are in $\mathcal{P}$, and $k_i$ is a real number ($i = 1...n$). When $\mathcal{C}$ contains only unary constraints, the DTP is called *Simple Temporal Problem* (STP) (Dechter, Meiri, & Pearl, 1991).

A DTP is *consistent* if and only if the DTP has a solution. A *solution* of a DTP is an assignment of real values to the variables of the DTP that is consistent with every constraint in the DTP. Computing a solution for a DTP is an NP-hard problem (Dechter et al., 1991), while computing a solution of an STP can be accomplished in polynomial time. Given an STP with a special "start time" variable $s$ preceding all the others, we can compute a solution of the STP where each variable has the shortest possible distance from $s$ in $O(n \cdot c)$ time, for $n$ variables and $c$ constraints in the STP (Dechter et al., 1991; Gerevini & Cristani, 1997). We call such a solution an *optimal solution* of the STP. Clearly, a DTP is consistent if and only if we can choose from each constraint in the DTP a disjunct obtaining a consistent STP, and any solution of such an STP is also a solution of the original DTP.

Finally, an STP is consistent if and only if the *distance graph* of the STP does not contain negative cycles (Dechter et al., 1991). The distance graph of an STP $\langle \mathcal{P}, \mathcal{C} \rangle$ is a directed labeled graph with a vertex labeled $p$ for each $p \in \mathcal{P}$, and with an edge from $v \in \mathcal{P}$ to $w \in \mathcal{P}$ labeled $k$ for each constraint $w - v \leq k \in \mathcal{C}$.

## 2.2 Augmenting the LA-graph with Disjunctive Temporal Constraints

Let $p$ be a timed precondition over a set $W(p)$ of time windows. In the following, $x^-$ and $x^+$ indicate the start time and end time of $x$, respectively, where $x$ is either a time window or an action. Moreover, $a_l$ indicates an action node at level $l$ of the LA-graph under consideration. For clarity of presentation, we will describe our techniques focusing on action preconditions that must hold during the whole execution of the action (except at the end point of the action), and on operator effects that hold at the end of the action execution, i.e., on PDDL conditions of type "`over all`", and PDDL effects of type "`at end`" (Fox & Long, 2003).[2]

In order to represent plans where actions have durations and time windows for their execution, we augment the ordering constraints of an LA-graph with (i) action *duration constraints* and (ii) action *scheduling constraints*. Duration constraints have form

$$a^+ - a^- = Dur(a),$$

where $Dur(a)$ denotes the duration of an action $a$ (for the special actions $a_{start}$ and $a_{end}$, we have $Dur(a_{start}) = Dur(a_{end}) = 0$, since $a_{start}^- = a_{start}^+$ and $a_{end}^- = a_{end}^+$). Duration constraints are supported by the representation proposed in a previous work (Gerevini et al., 2003), while the representation and treatment of scheduling constraints are a major contribution of this work.

---

2. Our methods and planner support all the types of operator condition and effect that can be specified in PDDL 2.1 and 2.2.

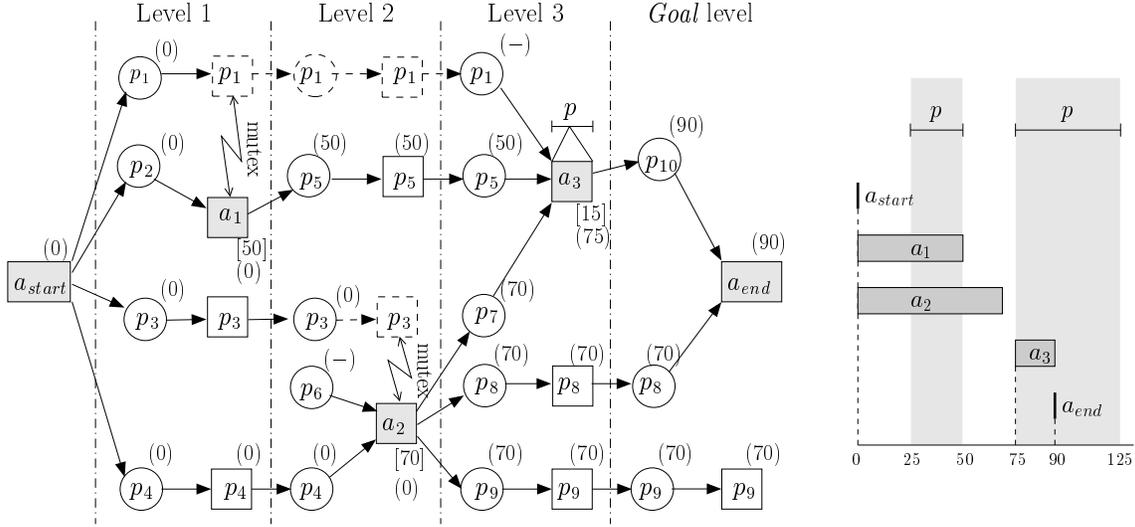

Figure 1: An example of LA-graph with nodes labeled by $\mathcal{T}$-values (in round brackets), and the Gantt chart of the actions labeling the nodes of the LA-graph. Square nodes are action nodes; circle nodes are fact nodes. Action nodes are also marked by the duration of the represented actions (in square brackets). Unsupported precondition nodes are labeled "(−)". Dashed edges form chains of no-ops blocked by mutex actions. Grey areas in the Gantt chart represent the time windows for the timed precondition $p$ of $a_3$.

Let $\pi$ be the plan represented by an LA-graph $\mathcal{A}$. It is easy to see that the set $\mathcal{C}$ formed by the ordering constraints in $\mathcal{A}$ and the duration constraints of the actions in $\pi$ can be encoded into an STP. For instance, if $a_i \in \pi$ is used to support a precondition node of $a_j$, then $a_i^+ - a_j^- \leq 0$ is in $\mathcal{C}$; if $a_i$ and $a_j$ are two mutex actions in $\pi$, and $a_i$ is ordered before $a_j$, then $a_i^+ - a_j^- \leq 0$ is in $\mathcal{C}$. Moreover, for every action $a \in \pi$, the following STP-constraints are in $\mathcal{C}$:

$$a^+ - a^- \leq Dur(a), \quad a^- - a^+ \leq -Dur(a),$$

which are equivalent to $a^+ - a^- = Dur(a)$. A scheduling constraint imposes the constraint that the execution of an action must occur during the time windows associated with a timed precondition of the action. Syntactically, it is a disjunctive constraint $c_1 \vee \cdots \vee c_n$, where $c_i$ is of the form

$$(y_i^\pm - x_i^\pm \leq h_i) \wedge (v_i^\pm - u_i^\pm \leq k_i),$$

$u_i^\pm, v_i^\pm, x_i^\pm, y_i^\pm$ are action start times or action end times, and $h_i, k_i \in \mathbb{R}$. For every action $a \in \pi$ with a timed precondition $p$, the following disjunctive constraint is added to $\mathcal{C}$:

$$\bigvee_{w \in W(p)} \left( \left( a_{start}^+ - a^- \leq -w^- \right) \wedge \left( a^+ - a_{start}^+ \leq w^+ \right) \right).^3$$

**Definition 1** *A **temporally disjunctive action graph** (TDA-graph) is a 4-tuple $\langle \mathcal{A}, \mathcal{T}, \mathcal{P}, \mathcal{C} \rangle$ where*

- $\mathcal{A}$ *is a linear action graph;*

- $\mathcal{T}$ *is an assignment of real values to the nodes of $\mathcal{A}$;*

- $\mathcal{P}$ *is the set of time point variables corresponding to the start times and the end times of the actions labeling the action nodes of $\mathcal{A}$;*

- $\mathcal{C}$ *is a set of ordering constraints, duration constraints and scheduling constraints involving variables in $\mathcal{P}$.*

A TDA-graph $\langle \mathcal{A}, \mathcal{T}, \mathcal{P}, \mathcal{C} \rangle$ represents the (partial) plan formed by the actions labeling the action nodes of $\mathcal{A}$ with start times assigned by $\mathcal{T}$. Figure 1 gives the LA-graph and $\mathcal{T}$-values of a simple TDA-graph containing five action nodes $(a_{start}, a_1, a_2, a_3, a_{end})$ and several fact nodes representing ten facts. The ordering constraints and duration constraints in $\mathcal{C}$ are:[4]

$$a_1^+ - a_3^- \leq 0, \ \ a_2^+ - a_3^- \leq 0,$$
$$a_1^+ - a_1^- = 50, \ \ a_2^+ - a_2^- = 70, \ \ a_3^+ - a_3^- = 15.$$

Assuming that $p$ is a timed precondition of $a_3$ with windows $[25, 50]$ and $[75, 125]$, the only scheduling constraint in $\mathcal{C}$ is:

$$((a_{start}^+ - a_3^- \leq -25) \wedge (a_3^+ - a_{start}^+ \leq 50)) \ \vee \ ((a_{start}^+ - a_3^- \leq -75) \wedge (a_3^+ - a_{start}^+ \leq 125)).$$

The pair $\langle \mathcal{P}, \mathcal{C} \rangle$ defines a DTP $\mathcal{D}$.[5] Let $\mathcal{D}_s$ be the set of scheduling constraints in $\mathcal{D}$. We have that $\mathcal{D}$ represents a set $\Theta$ of STPs, each of which consists of the constraints in $\mathcal{D} - \mathcal{D}_s$ and one disjunct (pair of STP-constraints) for each disjunction in a subset $\mathcal{D}_s'$ of $\mathcal{D}_s$ ($\mathcal{D}_s' \subseteq \mathcal{D}_s$). We call a consistent STP in $\Theta$ an *induced STP* of $\mathcal{D}$. When an induced STP contains a disjunct for every disjunction in $\mathcal{D}_s$ (i.e., $\mathcal{D}_s' = \mathcal{D}_s$), we say that such a (consistent) STP is a *complete induced STP* of $\mathcal{D}$.

The values assigned by $\mathcal{T}$ to the action nodes of $\mathcal{A}$ are the action start times corresponding to an optimal solution of an induced STP. We call these start times a *schedule* of the actions in $\mathcal{A}$. The $\mathcal{T}$ value labeling a fact node $f$ of $\mathcal{A}$ is the earliest time $t = \mathcal{T}_a + Dur(a)$ such that $a$ supports $f$ in $\mathcal{A}$, and $a$ starts at $\mathcal{T}_a$. If the induced STP from which we derive a schedule is incomplete, then $\mathcal{T}$ may violate the scheduling constraint of some action nodes, that we say are *unscheduled* in the current TDA-graph.

---

3. Note that, if $p$ is an `over all` timed condition of an action $a$, then the end of $a$ can be the time when an exogenous event making $p$ false happens, because in PDDL $p$ is not required to be true at the end of $a$ (Fox & Long, 2003).

4. For brevity, in our examples we omit the constraints $a_{start}^+ - a_i^- \leq 0$ and $a_i^+ - a_{end}^- \leq 0$, for each action $a_i$, as well as the duration constraints of $a_{start}$ and $a_{end}$, which have duration zero.

5. The disjunctive constraints in $\mathcal{C}$ are not exactly in DTP-form. However, it is easy to see that every disjunctive constraint in $\mathcal{C}$ can be translated into an equivalent conjunction of constraints in exact DTP-form. We use our more compact notation for clarity and efficiency reasons.

The following definitions present the notions of optimality for a complete induced STP and of optimal schedule, which will be used in the next section.

**Definition 2** *Given a DTP $\mathcal{D}$ with a point variable $p$, a complete induced STP of $\mathcal{D}$ is an* **optimal induced STP** *of $\mathcal{D}$ for $p$ iff it has a solution assigning to $p$ a value that is less than or equal to the value assigned to $p$ by every solution of every other complete induced STP of $\mathcal{D}$.*

**Definition 3** *Given a DTP $\mathcal{D}$ of a TDA-graph $\mathcal{G}$, an* **optimal schedule** *for the actions in $\mathcal{G}$ is an optimal solution of an optimal induced STP of $\mathcal{D}$ for $a_{end}^{-}$.*

Note that an optimal solution minimizes the makespan of the represented (possibly partial) plan. The DTP $\mathcal{D}$ of the previous example (Figure 1) has two induced STPs: one with no time window for $p$ ($\mathcal{S}_1$), and one including the pair of STP-constraints imposing the time window $[75, 125)$ to $p$ ($\mathcal{S}_2$). The STP obtained by imposing the time window $[25, 50)$ to $p$ is not an induced STP of the DTP, because it is not consistent. $\mathcal{S}_1$ is a partial induced STP of $\mathcal{D}$, while $\mathcal{S}_2$ is complete and optimal for the start time of $a_{end}$. The temporal values derived from the optimal solution of $\mathcal{S}_2$ that are assigned by $\mathcal{T}$ to the action nodes of the TDA-graph are: $a_{start}^{-} = a_{start}^{+} = 0$, $a_1^{-} = 0$, $a_2^{-} = 0$, $a_3^{-} = 75$, $a_{end}^{-} = a_{end}^{+} = 90$.

## 2.3 Solving the DTP of a TDA-graph

In general, computing a complete induced STP of a DTP (if it exists) is an NP-hard problem that can be solved by a backtracking algorithm (Stergiou & Koubarakis, 2000; Tsamardinos & Pollack, 2003). However, given the particular structure of the temporal constraints forming a TDA-graph, we show that this task can be accomplished in polynomial time with a backtrack-free algorithm. Moreover, the algorithm computes an optimal induced STP for $a_{end}^{-}$.

In the following, we assume that each time window for a timed precondition is no shorter than the duration of its action (otherwise, the time window should be removed from those available for this precondition and, if no time window remains, then the action cannot be used in any valid plan). Moreover, without loss of generality, we can assume that each action has at most *one* timed precondition. It is easy to see that we can always replace a set of `over all` timed conditions of an action $a$ with a single equivalent timed precondition, whose time windows are obtained by intersecting the windows forming the different original timed conditions of $a$. Also a set of `at start` timed conditions and a set of `at end` timed conditions can be compiled into single equivalent timed preconditions. This can be achieved by translating these conditions into conditions of type `over all`. The idea is similar to the one presented by Edelkamp (2004), with the difference that we can have more than one time window associated with a timed condition, while Edelkamp assumes that each timed condition is associated with a unique time window. Specifically, every `at start` timed condition $p$ of an action $a$ can be translated into an equivalent timed condition $p'$ of type `over all` by replacing the scheduling constraint of $p$,

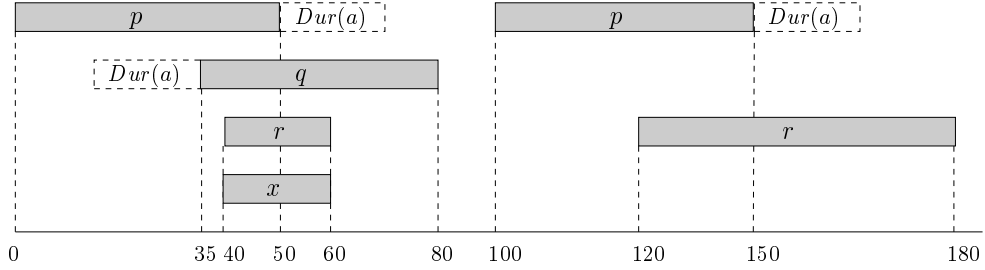

Figure 2: An example of a set of timed conditions compiled into a single timed precondition ($x$). The solid boxes represent the time windows associated with the timed conditions $p$ (of type `at start`), $q$ (of type `at end`), and $r$ (of type `over all`) of an action $a$. A solid box extended by a dashed box indicates the extension of the time window in the translation of the corresponding timed condition into an `over all` timed condition for $a$.

$$\bigvee_{w \in W(p)} \left( \left( a_{start}^+ - a^- < -w^- \right) \wedge \left( a^- - a_{start}^+ < w^+ \right) \right),$$

forcing $a^-$ to occur during one or more time windows, with

$$\bigvee_{w \in W(p)} \left( \left( a_{start}^+ - a^- < -w^- \right) \wedge \left( a^+ - a_{start}^+ < w^+ + Dur(a) \right) \right).[6]$$

Similarly, every `at end` timed condition $p$ can be translated into an equivalent `over all` timed condition by replacing the scheduling constraint

$$\bigvee_{w \in W(p)} \left( \left( a_{start}^+ - a^+ < -w^- \right) \wedge \left( a^+ - a_{start}^+ < w^+ \right) \right),$$

forcing $a^+$ to occur during one or more time windows, with

$$\bigvee_{w \in W(p)} \left( \left( a_{start}^+ - a^- < -w^- + Dur(a) \right) \wedge \left( a^+ - a_{start}^+ < w^+ \right) \right).$$

Clearly, this translation of the timed conditions of each domain action into a single timed precondition for the action can be accomplished by a preprocessing step in polynomial time. Figure 2 shows an example. Assume that action $a$ has duration 20 and timed conditions $p$ of type `at start`, $q$ of type `at end` and $r$ of type `over all`. Let $[0, 50)$ and $[100, 150)$ be the time windows of $p$, $[35, 80)$ the time window of $q$, and finally $[40, 60)$ and $[120, 180)$ the time windows of $r$. We can compile these timed conditions into a new timed condition $x$ with the time window $[40, 60)$.

---

6. Note that for timed conditions of type `at start` and `at end` we need to use "<" instead of "≤". However, the properties and algorithms for STPs can be easily generalized to STPs extended with <-constraints (e.g., Gerevini & Cristani, 1997).

Solve-DTP$(X, S)$

  *Input*: The set $X$ of meta-variables in the meta CSP of a DTP, a partial solution $S$ of the meta CSP;

  *Output*: Either a solution of the meta CSP or *fail*.

1.  **if** $X = \emptyset$ **then stop** and **return** $S$;
2.  $x \leftarrow$ SelectVariable$(X)$;  $X' \leftarrow X - \{x\}$;
3.  **while** $D(x) \neq \emptyset$ **do**
4.     $d \leftarrow$ SelectValue$(D(x))$;
5.     $S' \leftarrow S \cup \{x \leftarrow d\}$;  $D(x) \leftarrow D(x) - \{d\}$;
6.     $D'(x) \leftarrow D(x)$;   /* Saving the domain values */
7.     **if** ForwardCheck-DTP$(X', S')$ **then**
8.        Solve-DTP$(X', S')$;
9.     $D(x) \leftarrow D'(x)$;   /* Restoring the domain values */
10. **return** *fail*;   /* backtracking */

ForwardCheck-DTP$(X, S)$

  *Input*: The set $X$ of meta-variables, a (partial) solution $S$;

  *Output*: Either *true* or *false*.

1.  **forall** $x \in X$ **do**
2.     **forall** $d \in D(x)$ **do**
3.        **if** *not* Consistency-STP$(S \cup \{x \leftarrow d\})$ **then**
4.           $D(x) \leftarrow D(x) - \{d\}$;
5.     **if** $D(x) = \emptyset$ **then return** *false*;    /* dead-end */
6.  **return** *true*.

Figure 3:  Basic algorithm for solving a DTP. $D(x)$ is a global variable whose value is the current domain of the meta-variable $x$. Consistency-STP$(S)$ returns *true*, if the STP formed by the variable values in the (partial) solution $S$ has a solution, *false* otherwise.

As observed by Stergiou and Kourbarakis (2000) and Tsamardinos and Pollack (2003), a DTP can be seen as a "meta CSP": the variables of the meta CSP are the constraints of the original CSP, and the values of these (meta) variables are the disjuncts forming the constraints of the original CSP. The constraints of the meta CSP are not explicitly stated. Instead, they are implicitly defined as follows: an assignment $\theta$ of values to the meta-variables satisfies the constraints of the meta CSP iff $\theta$ forms a consistent STP (an induced STP of the DTP). A solution of the meta CSP is a complete induced STP of the DTP.

Figure 3 shows an algorithm for solving the meta CSP of a DTP (Tsamardinos & Pollack, 2003), which is a variant of the forward-checking backtracking algorithm for solving general CSPs. By appropriately choosing the next meta-variable to instantiate (function SelectVariable) and its value (function SelectValue), we can show that the algorithm finds a solution with *no backtracking* (if one exists). Moreover, by a simple modification of Solve-DTP, we can derive an algorithm that is backtrack free even when the input meta CSP has no solution. This can be achieved by exploiting the information in the LA-graph $\mathcal{A}$ of the TDA-graph to decompose its DTP $\mathcal{D}$ into a sequence of "growing DTPs"

$$\mathcal{D}_1 \subset \mathcal{D}_2 \subset ... \subset \mathcal{D}_{last} = \mathcal{D}$$

where (i) $last$ is the number of the levels in $\mathcal{A}$, (ii) the variables $V_i$ of $\mathcal{D}_i$ ($i = 1..last$) are all the variables of $\mathcal{D}$ corresponding to the action nodes in $\mathcal{A}$ up to level $i$, and (iii) the constraints of $\mathcal{D}_i$ are all the constraints of $\mathcal{D}$ involving only the variables in $V_i$. E.g., for the DTP of Figure 1, the point variables of $\mathcal{D}_3$ are $a^+_{start}$, $a^-_1$, $a^+_1$, $a^-_2$, $a^+_2$, $a^-_3$, $a^+_3$, and the set of constraints $\mathcal{D}_3$ is

$$\{a^+_1 - a^-_3 \leq 0,\ a^+_2 - a^-_3 \leq 0,\ a^+_1 - a^-_1 = 50,\ a^+_2 - a^-_2 = 70,\ a^+_3 - a^-_3 = 15,$$
$$((a^+_{start} - a^-_3 \leq -25) \wedge (a^-_3 - a^+_{start} \leq 50)) \vee ((a^+_{start} - a^-_3 \leq -75) \wedge (a^+_3 - a^+_{start} \leq 125))\}.$$

From the decomposed DTP, we can derive an ordered partition of the set of meta-variables in the meta CSP of the original DTP

$$X = X_1 \cup X_2 \cup ... \cup X_{last},$$

where $X_i$ is the set of the meta-variables corresponding to the constraints in $\mathcal{D}_i - \mathcal{D}_{i-1}$, if $i > 1$, and in $\mathcal{D}_1$ otherwise. This ordered partition is used to define the order in which SelectVariable chooses the next variable to instantiate, which is crucial to avoid backtracking. Specifically, every variable with a single domain value (i.e., an ordering constraint, a duration constraint, or a scheduling constraint with only one time window) is selected before every variable with more than one possible value (i.e., a scheduling constraint with more than one time window); moreover, if $x_i \in X_i$, $x_j \in X_j$ and $i < j$, then $x_i$ is selected before $x_j$.

In order to avoid backtracking, the order in which SelectValue chooses the value for a meta-variable is very important as well: given a meta-variable with more than one value (time window) in its current domain, we choose the value corresponding to the earliest available time window. E.g., if the current domain of the selected meta-variable with $m$ possible values is

$$\bigcup_{i=1..m} \left\{ \left(a^+_{start} - a^- \leq -w^-_i\right) \wedge \left(a^+ - a^+_{start} \leq w^+_i\right) \right\},$$

then SelectValue chooses the $j$-th value such that $|w^-_j| < |w^-_h|$, for every $h \in \{1, ..., m\}$, $j \in \{1, ..., m\}$, $h \neq j$.

In the following we give a simple example illustrating the order in which SelectVariable and SelectValue select the meta-variables and their meta-values, respectively. Consider the TDA-graph in Figure 1 with the additional time window $[150, 200]$ for the timed precondition $p$ of $a_3$. The DTP of the extended TDA-graph has six meta-variables ($x_1, x_2, \ldots, x_6$), whose domains (the disjuncts of the corresponding constraints of the original CSP) are:

$x_1$: $\{a^+_1 - a^-_3 \leq 0\}$
$x_2$: $\{a^+_2 - a^-_3 \leq 0\}$

$x_3: \{a_1^+ - a_1^- = 50\}$
$x_4: \{a_2^+ - a_2^- = 70\}$
$x_5: \{a_3^+ - a_3^- = 15\}$
$x_6: \{(a_{start}^+ - a_3^- \leq -25) \wedge (a_3^+ - a_{start}^- \leq 50), \ (a_{start}^+ - a_3^- \leq -75) \wedge (a_3^+ - a_{start}^- \leq 125),$
$\quad (a_{start}^+ - a_3^- \leq -150) \wedge (a_3^+ - a_{start}^+ \leq 200)\}.$

By exploiting the level structure of the TDA-graph, we derive an ordered partition of the meta-variables formed by the following sets:

$$X_1 = \{x_3\}, \ X_2 = \{x_4\}, \ X_3 = \{x_1, x_2, x_5, x_6\}.$$

Since $x_3$ belongs to $X_1$ while $x_4$ belongs to $X_2$, SelectVariable selects $x_3$ before selecting $x_4$. Similarly, the function selects $x_4$ before the meta-variables in $X_3$. When the algorithm instantiates $x_6$, the first meta-value of $x_6$ (i.e., the first time window of the timed precondition of $a_3$) has been removed from its domain by forward checking, and SelectValue selects $(a_{start}^+ - a_3^- \leq -75) \wedge (a_3^+ - a_{start}^+ \leq 125)$ before $(a_{start}^+ - a_3^- \leq -150) \wedge (a_3^+ - a_{start}^+ \leq 200)$, because the first meta-value corresponds to a time window starting at time 75, while the second one corresponds to a time window starting at time 150.

By using these techniques for selecting the next meta-variable to instantiate and its value, we can prove the following theorem.

**Theorem 1** *Given a DTP $\mathcal{D}$ for a TDA-graph, if the meta CSP $\mathcal{X}$ of $\mathcal{D}$ is solvable, then* Solve-DTP *finds a solution of $\mathcal{X}$ with no backtracking. Moreover, this solution is an optimal induced STP of $\mathcal{D}$ for $a_{end}^-$.*

**Proof.** The proof has two key points: the way meta-variables are selected and instantiated by SelectVariable and SelectValue, respectively; the particular type of constraints of $\mathcal{D}$, in which all disjunctive constraints have a specific form encoding a set of disjoint time windows, and, by construction of $\mathcal{D}$, we have

$$\forall j \ \neg \exists i \ such \ that \ i < j \ and \ \Omega \models a_j^{\pm} < a_i^{\pm}, \tag{1}$$

where $\Omega$ is the set of ordering constraints and duration constraints in $\mathcal{D}$, and $a_i^{\pm}$ ($a_j^{\pm}$) is an endpoint of $a_i$ ($a_j$). Because of property (1), $\Omega$ cannot imply any restriction on the *maximum* distance between an endpoint of $a_i$ and endpoint of $a_j$ (while, of course, there can be a lower bound on this distance). I.e., for any positive quantity $u$ we have

$$\forall j \ \neg \exists i \ such \ that \ i < j \ and \ \Omega \models (a_j^{\pm} - a_i^{\pm} \leq u). \tag{2}$$

Let assume that SelectVariable chooses a meta-variable $x$ that cannot be consistently instantiated to a value in $D(x)$ (and this means that we have reached a backtracking point). We show that this cannot be the case.

SelectVariable chooses the meta-variables of the STP-constraints of $\mathcal{D}$ before any meta-variable of a scheduling constraint with more than one value (time window). Let $X^s$ be the set of the meta-variables associated with the scheduling constraints in $\mathcal{D}$. We have that $x$ must be a meta-variable in $X^s$, because we are assuming that the meta CSP $\mathcal{X}$ is solvable. The use of the forward checking subroutine guarantees that at least one value of $x$ is consistent with respect to the meta-variables that are instantiated in the current partial solution $S$. Hence, it should be the case that at step 7 of Solve-DTP ForwardCheck-DTP returns *false* for every value $d$ (time window) in $D(x)$, i.e., that for every $d \in D(x)$ there exists another *un*instantiated meta-variable $x' \in X^s$ such that, for every $d' \in D(x')$, the check Consistency-STP$(S \cup \{x' \leftarrow d'\})$ executed by the forward checking subroutine returns *false*. However, if $\mathcal{X}$ has a solution ($\mathcal{D}$ is consistent), this cannot be the case because

(i) the value chosen by SelectValue to instantiate $x$ and the previously instantiated meta-variables (step 4) is the *earliest* available time window in the current domain of the meta-variable under consideration, which is a "least commitment assignment", and

(ii) we have at most one scheduling constraint (meta-variable in $X^s$) for each level of the TDA-graph.

Let $a'$ be the action constrained by the scheduling constraint associated with $x'$. Since SelectVariable selects $x$ before $x'$, by (ii) we have that $a'$ is at a level following the level of the action constrained by the scheduling constraint associated with $x$. Thus, by property (2), we have that if $x'$ could not be instantiated, then this would be because every time window of $a'$ constrains $a'$ to start "too early": the current partial solution of $\mathcal{X}$ augmented with any of the possible values of $x$ implies that the start time of $a'$ should be after the end of the last time window of $a'$. But then, (i) and the assumption that $\mathcal{X}$ is solvable guarantee that this cannot be the case.

Moreover, since the value of every instantiated meta-variable is propagated by forward checking to the unassigned variables, we have that the *first* value assigned to any meta-variable is the same value assigned to that variable in the solution found for the CSP (if any) – it is easy to see that if the first value chosen by SelectValue$(D(x))$ is not feasible (ForwardCheck-DTP$(X', S')$ returns false), then every other next value chosen for $x$ is not feasible.

Finally, since the value chosen by SelectValue for a meta-variable corresponds to the earliest available window in the current domain of the meta-variable, it follows that the solution computed by the algorithm is a complete optimal induced STP of $\mathcal{D}$ for $a^-_{end}$. □

As a consequence of the previous theorem, if Solve-DTP performs backtracking (step 10), then the input meta CSP has no solution. Thus, we can obtain a general backtrack-free algorithm for the DTP of any TDA-graph by simply replacing step 10 with

10. **stop** and **return** *fail*.

The correctness of the modified algorithm, which we called Solve-DTP$^+$, follows from Theorem 1. The next theorem states that the runtime complexity of Solve-DTP$^+$ is polynomial.

**Theorem 2** *Given a TDA-graph $\mathcal{G}$ with DTP $\mathcal{D}$, Solve-DTP$^+$ processes the meta CSP corresponding to $\mathcal{D}$ in polynomial time with respect to the number of action nodes in $\mathcal{G}$ and the maximum number of time windows in a scheduling constraint of $\mathcal{D}$.[7]*

---

7. It should be noted that here our main goal is to give a complexity bound that is polynomial. The use of improved forward checking techniques (e.g., Tsamardinos & Pollack, 2003) could lead to a complexity bound that is lower than the one given in the proof of the theorem.

**Proof.** The time complexity depends on the number of times ForwardCheck-DTP is executed, and on its time complexity. $\mathcal{D}$ contains a linear number of variables with respect to the number $n$ of domain action nodes in the LA-graph of the TDA-graph, $O(n^2)$ ordering constraints, and $O(n)$ duration constraints and scheduling constraints. Hence, the meta CSP of $\mathcal{D}$ has $O(n^2)$ meta-variables (one variable for each constraint of the original CSP). Let $\omega$ be the maximum number of time windows in a scheduling constraint of $\mathcal{D}$. ForwardCheck-DTP is executed at most $\omega$ times for each meta-variable $x$, i.e., $O(\omega \cdot n^2)$ times in total. Consistency-STP decides the satisfiability of an STP involving $O(n)$ variables, which can be accomplished in $O(n^3)$ time (Dechter et al., 1991; Gerevini & Cristani, 1997). (Note that the variables of the STP that is processed by Consistency-STP are the variables of the original CSP, i.e., they are the starting time and the end time of the actions in the plan.) Finally, Consistency-STP is run $O(\omega \cdot n^2)$ times during each run of ForwardCheck-DTP. It follows that the runtime complexity of Solve-DTP$^+$ is $O(\omega^2 \cdot n^7)$. □

By exploiting the structure of the temporal constraints forming the DTP of a TDA-graph, we can make the following additional changes to Solve-DTP$^+$ improving the efficiency of the algorithm.

- Instead of starting from an empty assignment $S$ (no meta-variable is instantiated), initially every meta-variable associated with an ordering constraint or with a duration constraint is instantiated with its value, and $X$ contains only meta-variables associated with the scheduling constraints. As observed in the proof of Theorem 1, if the meta CSP is solvable, the values assigned to the meta-variables by the initial $S$ form a consistent STP.

- Forward checking is performed only once for each meta-variable. This is because in the proof of Theorem 1 we have shown that, if the meta CSP is solvable, then the first value chosen by SelectValue should be feasible (i.e., ForwardCheck-DTP returns true). Thus, if the first value is not feasible, we can stop the algorithm and return *fail* because the meta CSP is not solvable. Moreover, we can omit steps 6 and 9 which save and restore the domain values of the meta-variables.

- Finally, the improved algorithm can be made incremental by exploiting the particular way in which we update the DTP of the TDA-graph during planning (i.e., during the search of a solution TDA-graph described in the next section). As described in the next section, each search step is either an addition of a new action node to a certain level $l$, or the removal of an action node from $l$. In both cases, it suffices to recompute the sub-solution for the meta-variables in the subsets $X_l, X_{l+1}, ..., X_{last}$. The values assigned to the other meta-variables is the same as the assignment in the last solution computed before updating the DTP, and it is part of the input of the algorithm.

Moreover, in order to use the local search techniques described in the next section, we need another change to the basic algorithm: when the algorithm detects that $\mathcal{X}$ has no solution, instead of returning failure, (i) it keeps processing the remaining meta-variables, and (ii) when it terminates, it returns the (partial) induced STP $\mathcal{S}_i$ formed by the values assigned to the meta-variables. The optimal solution of $\mathcal{S}_i$ defines the $\mathcal{T}$-assignment of the TDA-graph.

In the next section, $\mathcal{S}_{\mathcal{G}}$ denotes the induced STP for the DTP of a TDA-graph $\mathcal{G}$ computed by our method.

## 3. Local Search Techniques for TDA-Graphs

A TDA-graph $\langle \mathcal{A}, \mathcal{T}, \mathcal{P}, \mathcal{C} \rangle$ can contain two types of *flaw*: unsupported precondition nodes of $\mathcal{A}$, called *propositional flaws*, and action nodes of $\mathcal{A}$ that are not scheduled by $\mathcal{T}$, called *temporal flaws*. If a level of $\mathcal{A}$ contains a flaw, we say that this level is flawed. For example, if the only time window for $p$ in the TDA-graph of Figure 1 were $(25, 50)$, then level 3 would be flawed, because the start time of $a_3$ would be 70, which violates the scheduling constraint for $a_3$ imposing that this action must be executed during $(25, 50)$.

A TDA-graph with no flawed level represents a valid plan and is called a *solution graph*. In this section, we present new heuristics for finding a solution graph in a search space of TDA-graphs. These heuristics are used to guide a local search procedure, called Walkplan, that was originally proposed by Gerevini and Serina (1999) and that is the heart of the search engine of our planner.

The initial TDA-graph contains only $a_{start}$ and $a_{end}$. Each search step identifies the neighborhood $N(\mathcal{G})$ (successor states) of the current TDA-graph $\mathcal{G}$ (search state), which is a set of TDA-graphs obtained from $\mathcal{G}$ by adding a *helpful action node* to $\mathcal{A}$ or removing a *harmful action node* from $\mathcal{A}$ in an attempt to repair the *earliest* flawed level of $\mathcal{G}$.[8]

In the following, for the sake of brevity when we refer to an action node of a TDA-graph, we are implicitly referring to an action node of the LA-graph of a TDA-graph. Similarly for the level of a TDA-graph. Moreover, we remind the reader that $a_l$ denotes the action at level $l$, while $l_a$ denotes the level of action $a$.

**Definition 4** *Given a flawed level $l$ of a TDA-graph $\mathcal{G}$, an action node is* **helpful** *for $l$ iff its insertion into $\mathcal{G}$ at a level $i \leq l$ would remove a propositional flaw at $l$.*

**Definition 5** *Given a flawed level $l$ of a TDA-graph $\mathcal{G}$, an action node at a level $i \leq l$ is* **harmful** *for $l$ iff its removal from $\mathcal{G}$ would remove a propositional flaw at $l$, or would decrease the $\mathcal{T}$-value of $a_l$, if $a_l$ is unscheduled.*

### Examples of helpful action node and harmful action node

An action node representing an action with effect $p_1$ is helpful for level 3 of the TDA-graph of Figure 1 if it is added at level 2 or 3 (bear in mind that the insertion of an action node at level 3 determines an expansion of the TDA-graph postponing $a_3$ to level 4; more details are given at the end of the examples). Action node $a_3$ of Figure 1 is harmful for level 3, because its precondition node $p_1$ is unsupported; action node $a_1$ is harmful for level 3, because it blocks the no-op propagation of $p_1$ at level 1, which would support the precondition node $p_1$ at level 3. Moreover, assuming $W(p) = \{(25, 50)\}$, $a_3$ is unscheduled in the plan represented by the LA-graph. Action node $a_2$ is harmful for level 3, because the removal of $a_2$ from

---

$\mathcal{A}$ would decrease the temporal value of $a_3$. On the contrary, $a_1$ is not harmful for level 3, because its removal would not affect the possible scheduling of $a_3$. Notice that an action node can be both helpful and harmful: $a_3$ is harmful for level 3, and it is helpful for the goal level (because it supports the precondition node $p_{10}$ of $a_{end}$).

When we add an action node to a level $l$ that is not empty, the LA-graph is extended by one level, all action nodes from $l$ are shifted forward by one level (i.e., they are moved to their next level), and the new action is inserted at level $l$. Similarly, when we remove an action node from level $l$, the graph is "shrunk" by removing level $l$. Some additional details about this process are given in another paper (Gerevini et al., 2003). Moreover, as pointed out in the previous section, the addition (removal) of an action node $a$ requires us to update the DTP of $\mathcal{G}$ by adding (removing) the appropriate ordering constraints between $a$ and other actions in the LA-graph of $\mathcal{G}$, the duration constraint of $a$, and the scheduling constraint of $a$ (if any). From the updated DTP, we can use the method described in the previous section to revise $\mathcal{T}$, and to compute a possibly new schedule of the actions in $\mathcal{G}$ (i.e., an optimal solution of $\mathcal{S}_{\mathcal{G}}$).

The elements in $N(\mathcal{G})$ are evaluated using a *heuristic evaluation function $E$* consisting of two weighted terms, estimating their additional *search cost* and *temporal cost*, i.e., the number of search steps required to repair the new flaws introduced, and their contribution to the makespan of the represented plan, respectively. An element of $N(\mathcal{G})$ with the lowest combined cost is then selected using a "noise parameter" randomizing the search to escape from local minima (Gerevini et al., 2003). In addition, in order to escape local minima, the new version of our planner uses a short *tabu list* (Glover & Laguna, 1997). In the rest of this section, we will focus on the search cost term of $E$. The techniques that we use for the evaluation of the temporal cost and the (automatic) setting of the term weights of $E$ are similar to those that we introduced in a previous work (Gerevini et al., 2003).

The search cost of adding a helpful action node $a$ to repair a flawed level $l$ of $\mathcal{G}$ is estimated by constructing a *relaxed temporal plan $\pi$* achieving

(1) the unsupported precondition nodes of $a$, denoted by *Pre(a)*

(2) the propositional flaws remaining at $l$ after adding $a$, denoted by *Unsup(l)*, and

(3) the supported precondition nodes of other action nodes in $\mathcal{G}$ that would become *un*supported by adding $a$, denoted by *Threats(a)*.

Moreover, we estimate the number of additional temporal flaws that the addition of $a$ and $\pi$ to $\mathcal{G}$ would determine, i.e., we count the number of

(I) action nodes of $\mathcal{G}$ that would become *unscheduled* by adding $a$ and $\pi$ to $\mathcal{G}$,

(II) the unsatisfied timed preconditions of $a$, if $a$ is unscheduled in the TDA-graph extended with $a$ and $\pi$,

(III) the action nodes of $\pi$ with a scheduling constraint that we estimate cannot be satisfied in the context of $\pi$ and of $\mathcal{G}$.

The search cost of adding $a$ to $\mathcal{G}$ is the number of actions in $\pi$ plus (I), (II) and (III), which are new terms of the heuristic evaluation. Note that the action nodes of (I) are those that would have to be ordered after $a$ (because $a$ is used to achieve one of their preconditions, or these action nodes are mutex with $a$) and that, given the estimated end time of $\pi$ and the duration of $a$, would excessively increase their start time. In (II) we consider the original formulation of the timed preconditions of $a$ (i.e., the formulation before their possible compilation into one "merged" new precondition, as discussed in Section 2.3). Finally, to check the scheduling constraint of an action in $\pi$, we consider the estimated end time of the relaxed subplan of $\pi$ used to achieve the preconditions of this action.

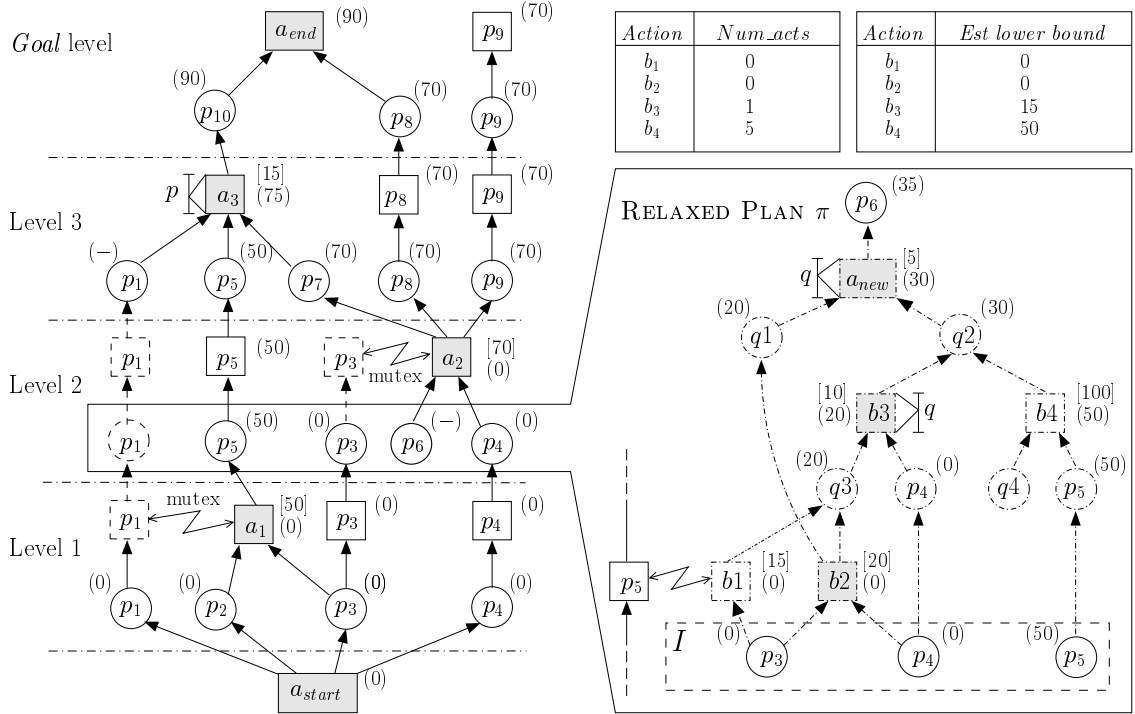

Figure 4: An example of relaxed temporal plan $\pi$. Square nodes represent action nodes, while the other nodes represent fact nodes; solid nodes correspond to nodes in $\mathcal{A} \cup \{a_{new}\}$; dotted nodes correspond to the precondition nodes and action nodes that are considered during the construction of $\pi$; the gray dotted nodes are those selected for their inclusion in $\pi$. Action nodes are marked by the duration of the represented actions (in square brackets) and by their estimated start time (in round brackets). The meaning of *Num_acts* is described in the text; the lower bounds on the earliest action start times (*Est lower bound*) are computed by the algorithm in Appendix A.

**Example of relaxed temporal plan and additional temporal flaws (I–III)**

Figure 4 gives an example of $\pi$ for evaluating the addition of $a_{new}$ at level 2 of the LA-graph on the left side of the figure (the same graph as the one used in Figure 1), which is a

RelaxedTimePlan$(G, I, A)$

  *Input*: A set of goal facts ($G$), an initial state for the relaxed plan ($I$), a set of reusable actions ($A$);

  *Output*: The set of actions $Acts$ forming a relaxed plan for $G$ from $I$ and the earliest time when all facts in $G$ can be achieved.

1.   $Acts \leftarrow A$;  $F \leftarrow \bigcup_{a \in Acts} Add(a)$;
2.   $t \leftarrow MAX \{T(g) \mid g \in G \cap F \text{ or } g \in G \cap I\}$;
3.   $G \leftarrow G - I$;
4.   **while** $G - F \neq \emptyset$
5.       $g \leftarrow$ a fact in $G - F$;
6.       $b \leftarrow BestAction(g)$;
7.       $\langle A, t' \rangle \leftarrow$ RelaxedTimePlan$(Pre(b), I, Acts)$;
8.       $T(b) \leftarrow ComputeEFT(b, t')$;
9.       $t \leftarrow MAX\{t, T(b)\}$;
10.      **forall** $f \in Add(b)$ **do**
11.          $T(f) \leftarrow MIN\{T(f), T(b) + Dur(b)\}$;
12.      $Acts \leftarrow A \cup \{b\}$;   $F \leftarrow \bigcup_{a \in Acts} Add(a)$;
13.  **return** $\langle Acts, t \rangle$.

Figure 5: Algorithm for computing a relaxed temporal plan. $ComputeEFT(b, t')$ returns the estimated earliest finishing time $\tau$ of $b$ that is consistent with the scheduling constraint of $b$ (if any), and such that $t' + Dur(b) \leq \tau$ (for an example see Appendix A). $Add(a)$ denotes the set of the positive effects of $a$.

helpful action node for the unsupported precondition $p_6$. The goals of $\pi$ are the unsupported preconditions $q1$ and $q2$ of $a_{new}$; while the initial state $I$ of $\pi$ is formed by the fact nodes that are supported at level 2. The actions of $\pi$ are $a_{new}$, $b2$ and $b3$. The numbers in the name of the actions and facts of the relaxed plan indicate the order in which RelaxedTimePlan considers them. The estimated start time and end time of $b3$ are 20 and 30, respectively. Assume that the timed precondition $q$ of $a_{new}$ has associated with it the time window $[0, 20)$. Concerning point (I), there is no action node of $\mathcal{G}$ that would become unscheduled by adding $a_{new}$ and $\pi$ to $\mathcal{G}$. Concerning point (II), $a_{new}$ is unscheduled and has one timed precondition that is unsatisfied ($q$). Concerning point (III), we have that $b3$ cannot be scheduled in the context of $\pi$ and the current TDA-graph $\mathcal{G}$. Finally, since $\pi$ contains three actions, and the sum of (I), (II) and (III) is 2, we have that the search cost of adding $a_{new}$ to $\mathcal{G}$ at level 2 is 5.

The evaluation of a TDA-graph derived by *removing* a harmful action node $a$ for a flawed level $l$ is similar, with $\pi$ achieving

- the precondition nodes supported by $a$ that would become *un*supported by removing $a$ and

- when $l_a < l$, the unsupported precondition nodes at level $l$ that do not become supported by removing $a$.

Regarding the second point, note that if $l = l_a$, then all flaws at $l$ are eliminated because, when we remove an action, we also (automatically) remove all its precondition nodes. While, when $l_a < l$, the removal of $a$ could leave some flaws at level $l$.

Plan $\pi$ is relaxed in the sense that its derivation ignores the possible (negative) interference between actions in $\pi$, and the actions in $\pi$ may be unscheduled. The derivation of $\pi$ takes into account the actions already in the current partial plan (the plan represented by the TDA-graph $\mathcal{G}$). In particular, the actions of the current plan are used to define an initial state $I$ for $\pi$, which is obtained by applying the actions of $\mathcal{G}$ up to level $l_a - 1$, ordered according to their corresponding levels. Moreover, each fact $f$ in $I$ is marked by a temporal value, $T(f)$, corresponding to the time when $f$ becomes true (and remains so in $\pi$) in the current subplan formed by the actions up to level $l_a - 1$.

The relaxed plan $\pi$ is constructed using a backward process, called RelaxedTimePlan (see Figure 5), which is an extension of the RelaxedPlan algorithm that we proposed in a previous work (Gerevini et al., 2003). The algorithm outputs two values: a set of actions forming a (sub)relaxed plan, and its estimated earliest finishing time (used to defined the temporal cost term of $E$). The set of actions $Acts$ forming $\pi$ is derived by running RelaxedTimePlan twice: first with goals $Pre(a)$, initial state $I$ and an empty set of reusable actions; then with goals $Unsup(l) \cup Threats(a)$, initial state $I - Threats(a) \cup Add(a)$, and a set of reusable actions formed by the actions computed by the first run plus $a$.

The main novelty of the extended algorithm for computing $\pi$ concerns the choice of the actions forming the relaxed plan. The action $b$ chosen to achieve a (sub)goal $g$ is an action minimizing the sum of

- the estimated minimum number of additional actions required to support its propositional preconditions from $I$ ($Num\_acts(b, I)$),

- the number of supported precondition nodes in the LA-graph that would become unsupported by adding $b$ to $\mathcal{G}$ ($Threats(b)$),

- the number of timed preconditions of $b$ that we estimate would be unsatisfied in $\mathcal{G}$ extended with $\pi$ ($TimedPre(b)$);

- the number of action nodes scheduled by $\mathcal{T}$ that we estimate would become unscheduled when adding $b$ to $\mathcal{G}$ ($TimeThreats(b)$).

More formally, the action chosen by $BestAction(g)$ at step 6 of RelaxedTimePlan to achieve a (sub)goal $g$ is an action satisfying

$$\underset{\{a' \in A_g\}}{ARGMIN}\Big\{ Num\_acts(a', I) + |\,Threats(a')\,| + |\,TimedPre(a')\,| + |\,TimeThreats(a')\,| \Big\},$$

where $A_g = \{a' \in \mathcal{O} \,|\, g \in Add(a'),$ $\mathcal{O}$ is the set of all the domain actions whose preconditions are reachable from $I\}$.

$Num\_acts(b, I)$ is computed by the algorithm given in Appendix A; $Threats(b)$ is computed as in our previous method for deriving $\pi$ (Gerevini et al., 2003), i.e., by considering the negative interactions (through mutex relations) of $b$ with the precondition nodes that are supported at levels after $a_l$; $TimedPre(b)$ and $TimeThreats(b)$ are new components of the action selection method, and they are computed as follows.

In order to compute $TimedPre(b)$, we estimate the earliest start time of $b$ ($Est(b)$) and the earliest finishing time of $b$ ($Eft(b)$). Using these values, we count the number of the timed preconditions of $b$ that cannot be satisfied. $Eft(b)$ is defined as $Est(b) + Dur(b)$, while $Est(b)$ is the maximum over

- a lower bound on the possible earliest start time of $b$ ($Est\ lower\ bound\ of\ b$), computed by the reachability analysis algorithm given in Appendix A;

- the $\mathcal{T}$-values of the action nodes $c_i$ in the current TDA-graph $\mathcal{G}$, with $i < l_a$, that are mutex with $b$ because the addition of $b$ to $\mathcal{G}$ would occur the addition of $c_i^+ - b^- \leq 0$ to the DTP of $\mathcal{G}$;

- the maximum over an estimated lower bound on the time when all the preconditions of $b$ are achieved in the relaxed plan; this estimate is computed from the causal structure of the relaxed plan, the duration and scheduling constraints of its actions, and the $\mathcal{T}$-values of the facts in the initial state $I$.

**Example of "TimedPre"**

In the example of Figure 4, the estimated start time of $b3$ is the maximum between 15, which is the $Est$ lower bound of $b3$, and 20, which is the maximum time over the estimated times when the preconditions of $b3$ are supported ($p_4$ is supported in the initial state of $\pi$ at time 0, while $q_3$ is supported at time 20). Notice that $a_1$ is not mutex with $b3$, and so the second point in the definition of $Est(b3)$ does not apply here. Since the estimated earliest start time of $b3$ is 20 and the duration of $b_3$ is 10, $Eft(b3) = 20 + 10$. Thus, if we assume that $q$ has associated with it the time window [0,20], then the timed precondition $q$ of $b3$ cannot be scheduled, i.e., $q \in TimedPre(b3)$.

In order to compute $TimeThreats(b)$, we use the following notion of *time slack* between action nodes.

**Definition 6** *Given two action nodes $a_i$ and $a_j$ of a TDA-graph $\langle \mathcal{A}, \mathcal{T}, \mathcal{P}, \mathcal{C} \rangle$ such that $\mathcal{C} \models a_i^+ < a_j^-$, $Slack(a_i, a_j)$ is the maximum time by which the $\mathcal{T}$-value of $a_i$ can be consistently increased in $\mathcal{S}_{\mathcal{G}}$ without violating the time window chosen for scheduling $a_j$.*

In order to estimate whether an action $b$ is a time threat for an action node $a_k$ in the current TDA-graph extended with the action node $a$ that we are adding for repairing level $l$ ($l < k$), we check if

$$\Delta(\pi_b, a) > Slack(a, a_k)$$

holds, where $\pi_b$ is the portion of the relaxed plan computed so far, and $\Delta(\pi_b, a)$ is the estimated delay that adding the actions in $\pi_b$ to $\mathcal{G}$ would cause to the start time of $a$.

**Examples of time slack and "TimeThreat"**

The slack between $a_{new}$ and $a_3$ in the TDA-graph of Figure 4 extended with $a_{new}$ is 35, because even if $a_{new}$ started at 35, $a_3$ could still be executed during the time window $[75, 125]$ (imposed by the timed precondition $p$); while if $a_{new}$ started at $35 + \epsilon$, then $a_3$ would finish at $125 + \epsilon$ (determined by summing the start time of $a_{new}$, $Dur(a_{new})$, $Dur(a_2)$, and $Dur(a_3)$), and so the scheduling constraint of $a_3$ would be violated. Assume that we are evaluating the inclusion of $b4$ in the relaxed plan of Figure 4 for achieving $q2$. We have

$$\Delta(\pi_{b4}, a_{new}) = 150,$$

i.e. the estimated delay that the portion of the plan formed by $b4$ would add to the end time of $a_{new}$ is 150. Since the slack between $a_{new}$ and $a_3$ is 35,

$$Slack(a_{new}, a_3) < \Delta(\pi_{b4}, a_{new}),$$

and so $a_3 \in TimeThreats(b4)$. On the contrary, since

$$Slack(a_{new}, a_3) > \Delta(\pi_{b3}, a_{new}) = 30$$

we have that $a_3 \notin TimeThreats(b3)$.

To conclude this section, we observe that the way we consider scheduling constraints during the evaluation of the search neighborhood has some similarity with a well-known technique used in scheduling. For example, suppose that we are evaluating the TDA-graphs obtained by adding a helpful action node $a$ to one among some alternative possible levels of the graph, and that the current TDA-graph contains another action node $c$ which is mutex with $a$. If the search neighborhood contains two TDA-graphs corresponding to (1) "adding $a$ to a level before $l_c$" and (2) "adding $a$ to a level after $l_c$", and (1) violates less scheduling constraints than (2), then, according to points (I)–(III), (1) is preferred to (2). A similar heuristic method, called *constraint-based analysis*, has been proposed by Erschler, Roubellat and Vernhes (1976) to decide whether an action should be scheduled before or after another conflicting action, and it has been also used in other scheduling work for guiding the search toward a consistent scheduling of the tasks involved in the problem (e.g., Smith & Cheng, 1993).

## 4. Experimental Results

We implemented our approach in a planner called LPG-td, which obtained the 2nd prize in the metric-temporal track ("satisficing planners") of the 4th International Planning Competition (IPC-4). LPG-td is an incremental planner, in the sense that it produces a sequence of valid plans each of which improves the quality of the previous ones. Plan quality is measured using the metric expression that is specified in the planning problem description. The incremental process of LPG-td is described in another paper (Gerevini et al., 2003). Essentially, the process iterates the search of a solution graph with an additional constraint on the lower bound of the plan quality, which is determined by the quality of the previously generated plans. LPG-td is written in C and is available from `http://lpg.ing.unibs.it`.

In this section, we present the results of an experimental study with two main goals:

- testing the efficiency of our approach to temporal planning with predictable exogenous events by comparing the performance of LPG-td and other recent planners that at IPC-4 attempted the benchmark problems involving timed initial literals (Edelkamp, Hoffmann, Littman, & Younes, 2004);

| Planner | Solved | Attempted | Success ratio | Planning capabilities at IPC-4 |
|---------|--------|-----------|---------------|-------------------------------|
| LPG-td | 845 | 1074 | 79% | Propositional + DP, Metric-Temporal +TIL |
| SGPLAN | 1090 | 1415 | 77% | Propositional + DP, Metric-Temporal +TIL |
| P-MEP | 98 | 588 | 17% | Propositional, Metric-Temporal +TIL |
| CRIKEY | 364 | 594 | 61% | Propositional, Metric-Temporal |
| LPG-IPC3 | 306 | 594 | 52% | Propositional, Metric-Temporal |
| DOWNWARD (diag) | 380 | 432 | 88% | Propositional + DP |
| DOWNWARD | 360 | 432 | 83% | Propositional + DP |
| MARVIN | 224 | 432 | 52% | Propositional + DP |
| YAHSP | 255 | 279 | 91% | Propositional |
| MACRO-FF | 189 | 332 | 57% | Propositional |
| FAP | 81 | 193 | 42% | Propositional |
| ROADMAPPER | 52 | 186 | 28% | Propositional |
| TILSAPA | 63 | 166 | 38% | TIL |
| OPTOP | 4 | 50 | 8% | TIL |

Table 1: Number of problems attempted/solved and success ratio of the (satisficing) planners that took part in IPC-4. "DP" means derived predicates; "TIL" means timed initial literals; "Propositional" means STRIPS or ADL. The planning capabilities are the PDDL2.2 features in the test problems attempted by each planner at IPC-4.

- testing the effectiveness of the proposed temporal reasoning techniques integrated into the planning process to understand, in particular, their impact on the overall performance of the system, and to compare them with other existing techniques.

For the first analysis, we consider the test problems of the variant of the IPC-4 metric-temporal domains involving timed initial literals. A comparison of LPG-td and other IPC-4 planners considering all the variants of the IPC-4 metric-temporal domains is given in Appendix B. Additional results are available from the web site of our planner.

For the second experiments, we use new domains and problems obtained by extending two well-known benchmark domains (and the relative problems) from IPC-3 with timed initial literals (Long & Fox, 2003a).[9]

All tests were conducted on an Intel Xeon(tm) 3 GHz, 1 Gbytes of RAM. We ran LPG-td with the same default settings for every problem attempted.

## 4.1 LPG-td and Other IPC-4 Planners

In this section, we use the official results of IPC-4 to compare the performance of LPG-td with those of other planners that took part in the competition. The performance of LPG-td corresponds to a single run. The CPU-time limit for the run was 30 minutes, after which termination was forced. LPG-td.s indicates the CPU-time required by our planner to derive the first plan; LPG-td.bq indicates the best quality plan found within the CPU-time limit.

---

9. For a description of the IPC-4 domains and of the relative variants, the reader can visit the official web site of IPC-4 (http://ls5-www.cs.uni-dortmund.de/~edelkamp/ipc-4/index.html). The extended versions of the IPC-3 domains used in our experiments are available from http://zeus.ing.unibs.it/lpg/TestsIPC3-TIL.tgz.

Before focusing our analysis on the IPC-4 domains involving timed initial literals, in Table 1 we give a very brief overview of all the results of the IPC-4 (satisficing) planners, in terms of planning capabilities and problems attempted/solved by each planner. The table summarizes the results for all the domain variants of IPC-4. LPG-td and SGPLAN (Chen, Hsu, & Wah B., 2004) are the only planners supporting all the major features of PDDL2.1 and PDDL2.2. Both planners have a good success ratio (close to 80%). DOWNWARD (Helmert, 2004) and YAHSP (Vidal, 2004) have a success ratio better than LPG-td and SGPLAN, but they handle only propositional domains (DOWNWARD supports derived predicates, while YAHSP does not). SGPLAN attempted more problems than LPG-td because it was also tested on the "compiled version" of the variants with derived predicates and timed initial literals.[10] Moreover, LPG-td did not attempt the numerical variant of the two versions of the `Promela` domain and the ADL variant of `PSR-large`, because they use equality in some numerical preconditions or conditional effects, which currently our planner does not support.

Figure 6 shows the performance of LPG-td in the variants of three domains involving predictable exogenous events with respect to the other (satisficing) planners of IPC-4 supporting timed initial literals: SGPLAN, P-MEP (Sanchez et al., 2004) and TILSAPA (Kavuluri & U, 2004). In `Airport` (upper plots of the figure), LPG-td solves 45 problems over 50, SGPLAN 43, P-MEP 12, and TILSAPA 7. In terms of CPU-time, LPG-td performs much better than P-MEP and TILSAPA. LPG-td is faster than SGPLAN in nearly all problems (except problems 1 and 43). In particular, the gap of performance in problems 21–31 is nearly one order of magnitude. Regarding plan quality, the performance of LPG-td is similar to the performance of P-MEP and TILSAPA, while, overall, SGPLAN finds plan of worse quality (with the exception of problems 41 and 43, where SGPLAN performs slightly better, and the easiest problems where LPG-td and SGPLAN perform similarly).

LPG-td and TILSAPA are the only planners of IPC-4 that attempted the variant of `PipesWorld` with timed initial literals (central plots of Figure 6). LPG-td solves 23 problems over 30, while TILSAPA solves only 3 problems. In this domain variant LPG-td performs much better than TILSAPA.

In the "flaw version" of `Umts` (bottom plots of Figure 6), LPG-td solves all 50 problems, while SGPLAN solves 27 problems (P-MEP and TILSAPA did not attempt this domain variant). Moreover, LPG-td is about one order of magnitude faster than SGPLAN in every problem solved. Compared to the other IPC-4 benchmark problems, the `Umts` problems are generally easier to solve. In these test problems, the main challenge is finding plans of good quality. Overall, the best quality plans of LPG-td are much better than SGPLAN plans, except for the simplest problems where the two planners generate plans of similar quality. In the basic version of `Umts` without flawed actions, SGPLAN solves all problems as LPG-td, but in terms of plan quality LPG-td performs much better.

Figure 7 shows the results of the Wilcoxon sign-rank test, also known as the "Wilcoxon matched pairs test" (Wilcoxon & Wilcox, 1964), comparing the performance of LPG-td and the planners that attempted the benchmark problems of IPC-4 involving timed initial literals. The same test has been used by Long and Fox (2003a) for comparing the performance of the IPC-3 planners. For the CPU-time analysis, we consider all the problems *attempted* by both the compared planners and solved by at least one of them (when a planner does not solve a problem, the corresponding CPU-time is the IEEE arithmetic representation of positive infinity). For the plan quality (makespan) analysis, we consider all the problems *solved* by both the compared planners.

---

10. Such versions were generated for planners that do not support these features of PDDL2.2. During the competition we did not test LPG-td with the problems of the compiled domains because the planner supports the original version of these domains. LPG-td attempted every problem of the (uncompiled) IPC-4 domains that it could attempt in terms of the planning language it supports.

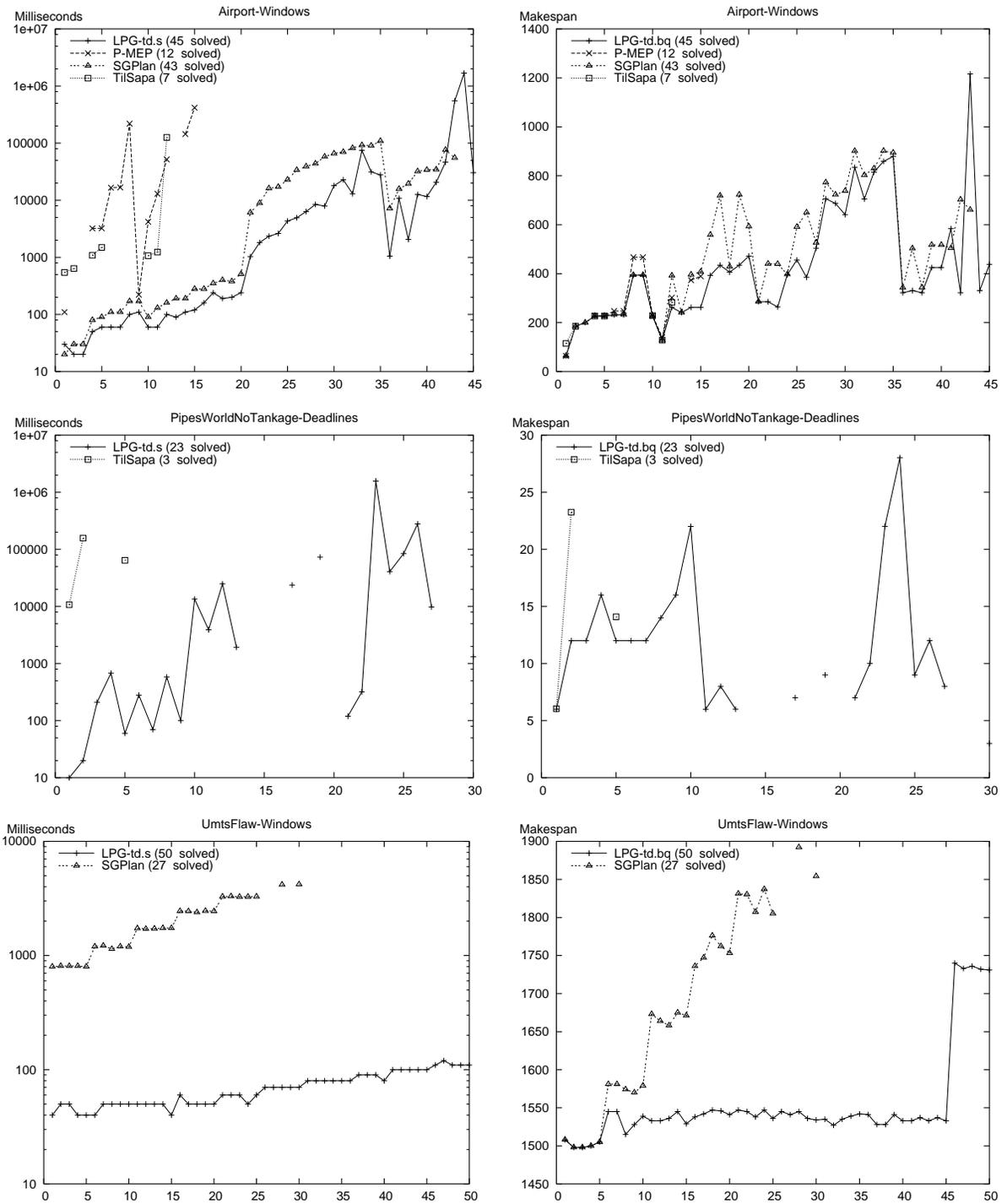

Figure 6: CPU-time and plan quality of LPG-td, P-MEP, SGPLAN, and TILSAPA for three IPC-4 domains with timed initial literals. On the *x*-axis we have the problem names simplified by numbers. In the plots on the left, on the y-axis we have CPU-milliseconds (logarithmic scale); in the plots on the right, on the *y*-axis we have the plan makespan (the lower the better).

| CPU-Time Analysis | | |
|---|---|---|
| LPG-td.s *vs* P-MEP | LPG-td.s *vs* SGPLAN | LPG-td.s *vs* TILSAPA |
| 5.841 | 3.162 | 10.118 |
| < 0.001 | (0.0016) | < 0.001 |
| 45 | 197 | 136 |

| Plan Quality Analysis | | |
|---|---|---|
| LPG-td.bq *vs* P-MEP | LPG-td.bq *vs* SGPLAN | LPG-td.bq *vs* TILSAPA |
| | 9.837 | 6.901 |
| < 0.001 | < 0.001 | < 0.001 |
| 12 | 154 | 63 |

Figure 7: Results of the Wilcoxon test for the performance of LPG-td compared with other IPC-4 (satisficing) planners in terms of CPU-times and plan quality for the benchmark problems with timed initial literals.

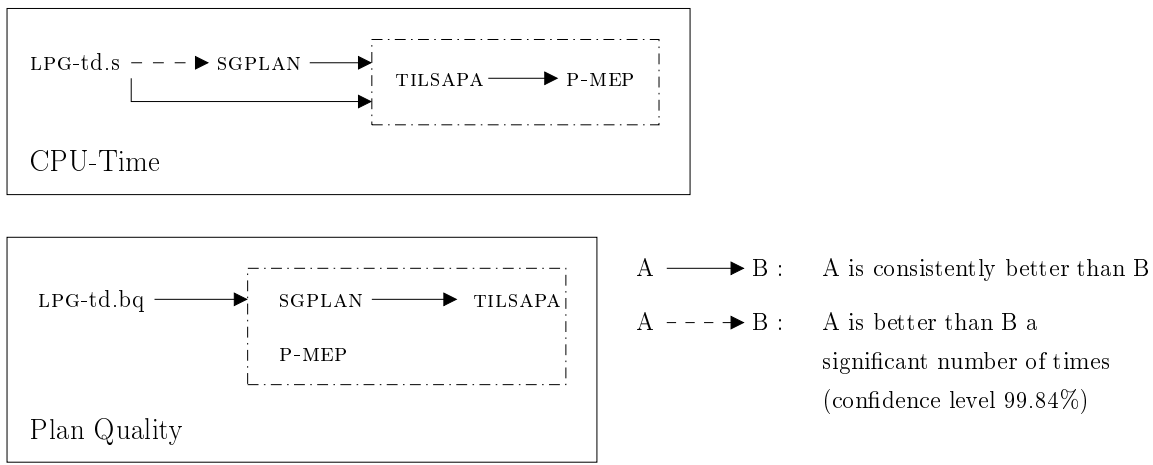

Figure 8: Partial order of the performance of the IPC-4 (satisficing) planners according to the Wilcoxon test for the benchmark problems with timed initial literals. A dashed arrow indicates that the performance relationship holds with a confidence level slightly less than 99.9%.

In order to carry out the Wilcoxon test, for each planning problem we computed the difference between the CPU-times of the two planners being compared, defining the samples of the test for the CPU-time analysis. Similarly, for the test concerning the plan quality analysis we computed the differences between the makespan of the plans generated by the two planners. The absolute values of these differences are ranked by increasing numbers starting from the lowest value. (The lowest value is ranked 1, the next lowest value is ranked 2, and so on.) Then we sum the ranks of the positive differences, and we sum the ranks of the negative differences. If the performance of the two planners is not significantly different, then the number of the positive differences will be approximately equal to the number of the negative differences, and the sum of the ranks in the set of the positive differences will be approximately equal to the sum of the ranks in the other set. Intuitively, the test considers a weighted sum of the number of times one planner performs better than the other. The sum is weighted because the test uses the performance gap to assign a rank to each performance difference.

Each cell in Figure 7 gives the result of a comparison between the performance of LPG-td and another IPC-4 planner. When the number of the samples is sufficiently large, the T-distribution used by the Wilcoxon test is approximatively a normal distribution. Therefore, the cells of the figure contain the $z$-value and the $p$-value characterizing the normal distribution. The higher the $z$-value, the more significant the difference of the performance is. The $p$-value represents the level of significance in the performance gap. We use a confidence level of 99.9%; hence, if the $p$-value is lower than 0.001, then the performance of the compared planners is statistically different. When this information appears on the left (right) side of the cell, the first (second) planner named in the title of the cell performs better than the other planner.[11] For the analysis comparing the CPU-time, the value under each cell is the number of the problems solved by at least one planner; while for the analysis comparing the plan quality, it is the number of problems solved by both the planners.

Figure 8 shows a graphical description of the relative performance of the IPC-4 satisficing planners according to the Wilcoxon test for the benchmark problems with timed initial literals. A solid arrow from a planner A to a planner B (or a cluster of planners B) indicates that the performance of A is statistically different from the performance of B, and that A performs better than B (every planner in B). A dashed arrow from A to B indicates that A is better than B a significant number of times, but there is no significant Wilcoxon relationship between A and B with a confidence level of 99.9% (on the other hand, the relationship holds with a confidence level slightly less than 99.9%). The results of this analysis say that LPG-td is consistently faster than TILSAPA and P-MEP, while it is faster than SGPLAN a significant number of times. In terms of plan quality, LPG-td performs consistently better than P-MEP, SGPLAN and TILSAPA.

Although LPG-td does not guarantee optimal plans, it is interesting to compare its performance with the optimal planners that took part in IPC-4, especially to see how good LPG-td's plans are. Figure 9 shows the performance of LPG-td and the best results over the results of *all* the other optimal IPC-4 planners ("AllOthers-Opt") in the temporal variants of Airport and Umts (without flawed actions). The plots for the plan quality (makespan) show that, in nearly every problem of these domains, the best quality plan found by LPG-td is an optimal solution, and that the first plan found by LPG-td is generally a good solution. The plots for the CPU-time show that LPG-td finds a plan much more quickly than any optimal planner, and that the CPU-time required by LPG-td to find the best plan is often lower than the CPU-time required by AllOthers-Opt (except for problems 12, 16, 18 and 20 of `Airport`). It should be noted that LPG-td.bq is the last plan over a sequence of computed plans with increasing quality (and CPU-time). The intermediate plans in this sequence could already have good quality. In particular, as shown by the plan quality plot for `Airport`, the first plan (LPG-td.s) solving problem 12 has near-optimal quality, but it is computed much more quickly than the LPG-td.bq plan and the AllOthers-Opt plan.

---

11. The $p$-value in the cell comparing LPG-td and P-MEP is omitted because the number of problems solved by both LPG-td and P-MEP is not high enough to approximate the T-distribution to a normal distribution.

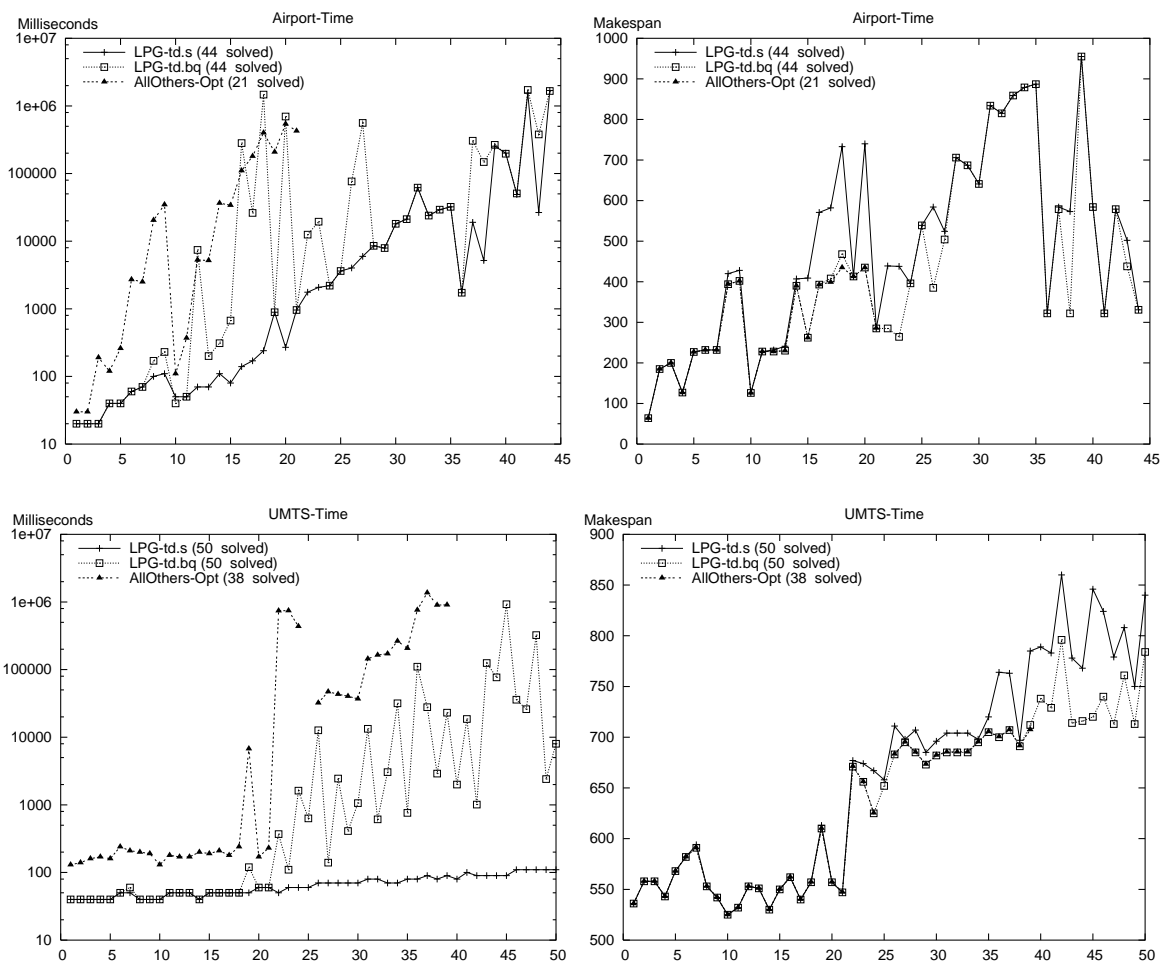

Figure 9: Performance of LPG-td and the best over all the optimal planners of IPC-4 (AllOthers-Opt) in `Airport-Time` and `Umts-Time`: CPU-time in logarithmic scale (left plots) and plan makespan (right plots). On the $x$-axis we have the problem names simplified by numbers.

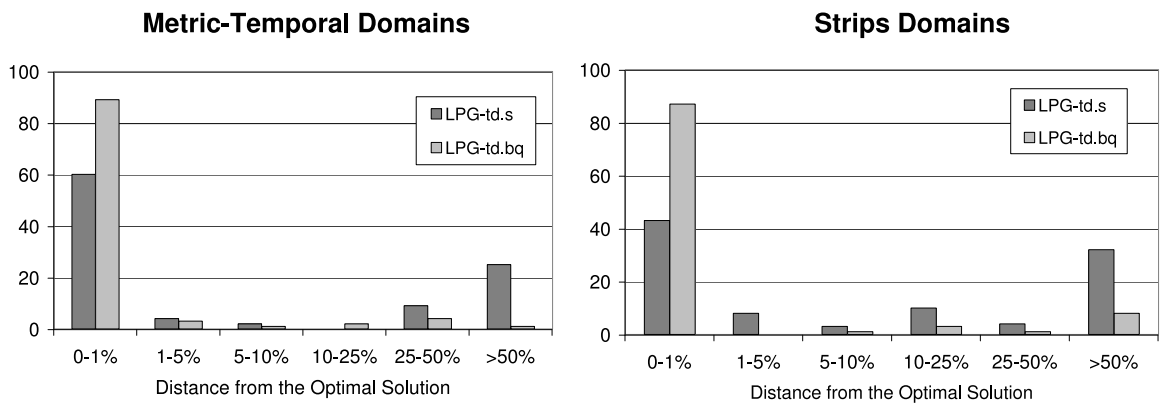

Figure 10: Plan quality distance between the solutions found by LPG-td and the corresponding optimal solutions. On the $x$-axis, we have some classes of quality distance (e.g., "10–25%" means that the plan generated by LPG-td is worse than the optimal plan by a factor between 0.1 and 0.25). On the $y$-axis, we have the percentage of solved problems for each of these classes.

Finally, Figure 10 gives the results of a more general analysis on the plan quality distance, considering all metric-temporal and STRIPS variants of the IPC-4 domains.[12] The analysis uses only the problems solved by at least one IPC-4 optimal planner. It is also important to note that we consider only the plans generated by the incremental process of LPG-td using no more CPU-time than the CPU-time required by the fastest optimal planner (AllOthers-Opt). Overall, the results in Figure 10 provide significant empirical evidence supporting the claim that often an incremental local search approach allows us to compute plans with very good quality using less or no more CPU-time than an optimal approach. In particular, the bars for the "0%–1%" class in the plot of the metric-temporal problems show that the percentage of the test problems in which the best quality plan of LPG-td (LPG-td.bq) is optimal or nearly optimal (i.e., plan quality is worse than optimal by a factor between 0 and 0.01, with 0 meaning no difference) is about 90%. Moreover, often the first plan computed by LPG-td (LPG-td.s) has good quality: 60% of all these plans have quality that is optimal or nearly optimal, and only about 25% of them have a quality that is worse than the optimal by a factor greater than 0.5.

Interestingly, the plot on the right of Figure 10 shows similar results concerning the good quality of LPG-td's plans also for the STRIPS problems of IPC-4 (with a lower percentage of the LPG-td.s' plans that are in the 0%–1% class, and a slightly higher percentage of the LPG-td.bq's plans that are in the "> 50%" class).

## 4.2 Temporal Reasoning in LPG-td

We conducted two main experiments. The first was aimed at testing the performance of LPG-td when the number of windows for the timed initial literals varies in problems having the same initial state and goals. The second experiment focused on our temporal reasoning techniques with the main goals of empirically evaluating their performance, and understanding their impact on the overall performance of LPG-td.

---

12. For the STRIPS problems, the plan quality metric is the number of the actions in the plan.

For these experiments we used two well-known IPC-3 domains, which were modified to include timed initial literals: `Rovers` and `ZenoTravel`. The version of `Rovers` with timed initial literals was obtained from the IPC-3 temporal version as follows. In the problem specification, for each "waypoint", we added a collection of pairs of timed initial literals of the type

```
(at t₁ (in_sun waypoint0))
(at t₂ (not (in_sun waypoint0)))
```

where $t_1 < t_2$. Each of these pairs defines a time window for the involved literal. In the operator specification file, the "`recharge`" operator has the precondition

```
(over all (in_sun ?w))
```

which imposes the constraint that the recharging actions are applied only when the rover is in the sun (`?w` is the operator parameter representing the waypoint of the recharging action.) The modified version of `ZenoTravel` was obtained similarly. In the problem specification, for each city we added a collection of pairs of timed initial literals of the type

```
(at t₁ (open-station city0))
(at t₂ (not (open-station city0)))
```

and in the operator specification file, we added the timed precondition

```
(over all (open-station ?c))
```

to the "`refuel`" operator, where `?c` is the operator parameter representing the city where the refuel action is executed.

Given a planning problem $\Pi$ and a collection of time windows $W_\phi$ for a timed literal $\phi$, it should be noted that, in general, the difficulty of solving $\Pi$ is affected by three parameters: the number of windows in $W_\phi$, their size, and the way they are distributed on the time line.[13] We considered two methods for generating test problems taking account of these parameters ($\Pi$ indicates an original IPC-3 problem in either the `Rovers` or `ZenoTravel` domain, and $n$ indicates the number of windows in $W_\phi$):

(I) Let $\pi$ be the best (shortest makespan) plan among those generated by LPG-td for solving $\Pi$ within a certain CPU-time limit, and $t$ the makespan of $\pi$. The time interval $[0, t]$ is divided into $2n - 1$ sub-intervals of equal size. The time windows for each timed literal $\phi$ of the extended problem $\Pi'$ are the odd "sub-intervals" of $[0, t]$, i.e.,

$$W_\phi = \left\{ \left[0, \frac{t}{2n-1}\right), \left[\frac{2t}{2n-1}, \frac{3t}{2n-1}\right), \dots, \left[\frac{(2n-2)t}{2n-1}, t\right] \right\}.$$

(II) Let $d$ be the maximum duration of an action in $\Pi$ with a timed precondition $\phi$. The time interval $\theta = [0, d \times (2n - 1)]$ is divided into $2n - 1$ sub-intervals of duration $d$.

---

13. In general, these parameters influence not only the hardness of temporal reasoning during planning, but also the logical part of the planning process (i.e., the selection of the actions forming the plan, that in LPG-td is done using heuristics taking exogenous events into account).

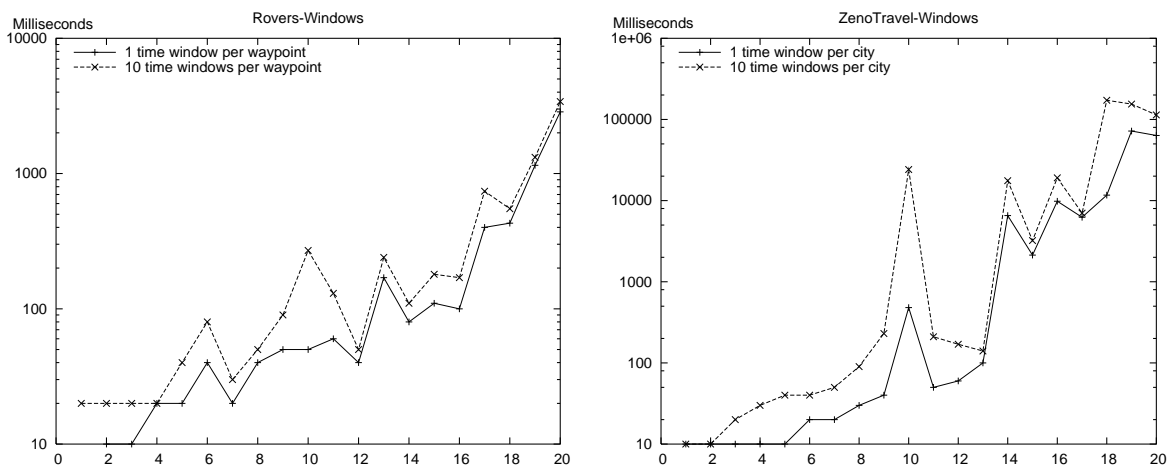

Figure 11: Performance of LPG-td in the `Rovers` and `ZenoTravel` domains extended with timed initial literals (1 and 10 time windows for each timed literal). The test problems were generated using method I. On the $x$-axis we have the problem names simplified by numbers; on the $y$-axis we have CPU-milliseconds (logarithmic scale).

Similarly to method (I), the time windows for $\phi$ in the extended problem $\Pi'$ are the odd sub-intervals of $\theta$.

Notice that we can use the first method only when the number of windows is relatively small because, if there are too many time windows of small size, the extended problem can become unsolvable (no window is large enough to schedule into it a necessary action with a timed precondition). The second method was designed to avoid this problem, and it can be used to test our techniques on planning problems involving many time windows.

Figures 11 and 12 give the results of the first experiment. The CPU-times in these plots are median values over five runs for each problem. For the results of Figure 11, we use the IPC-3 test problems modified by method I, while for the results of Figure 12 we use the IPC-3 test problems modified by method II. In both cases LPG-td solves all problems. The plots of Figure 11 indicate that the performance degradation when the number of windows increases from 1 to 10 is generally moderate, except in two cases. The plots of Figure 12 indicate that, when the number of windows increases exponentially from 1 to 10,000, the approach scales up well for the benchmark problems considered. For instance, consider the first `ZenoTravel` problem. With 1 window LPG-td solves this problem in 10 milliseconds, with 10 windows in 20 milliseconds, with 100 windows in 30 milliseconds, with 1000 windows in about 100 milliseconds, and with 10,000 windows in about 1 second. Moreover, we observed that the performance degradation is mainly determined by a heavier pre-processing phase (parsing and instantiation of the operators).

Tables 2 and 3 give some results concerning the experiment about our temporal reasoning techniques implemented in LPG-td. We consider some of the problems with 10 time windows (for each timed fluent) used for the tests of Figure 11, and we examine the computational cost of temporal reasoning during planning for these problems. In our approach to temporal planning, each search step defines a set of temporal constraints formed by the ordering and scheduling constraints in the current TDA-graph. Table 2 gives statistical information about such DTPs using both the compact constraint representation of LPG-td and the "classical" DTP representation. For each action in the TDA-graph, we have two temporal variables (the start/end times of the action), except $a_{start}$ and $a_{end}$ (for which, as we have pointed out, we can use only one variable). The number of the scheduling constraints and the number of the ordering constraints depend on which actions are in the current TDA-graph, and on how these actions are (causally or exclusively) related to each other, respectively (we have one scheduling constraint for each action with a timed precondition in the TDA-graph). Notice that our representation of the scheduling constraints is much more compact than the classical DTP formulation.[14]

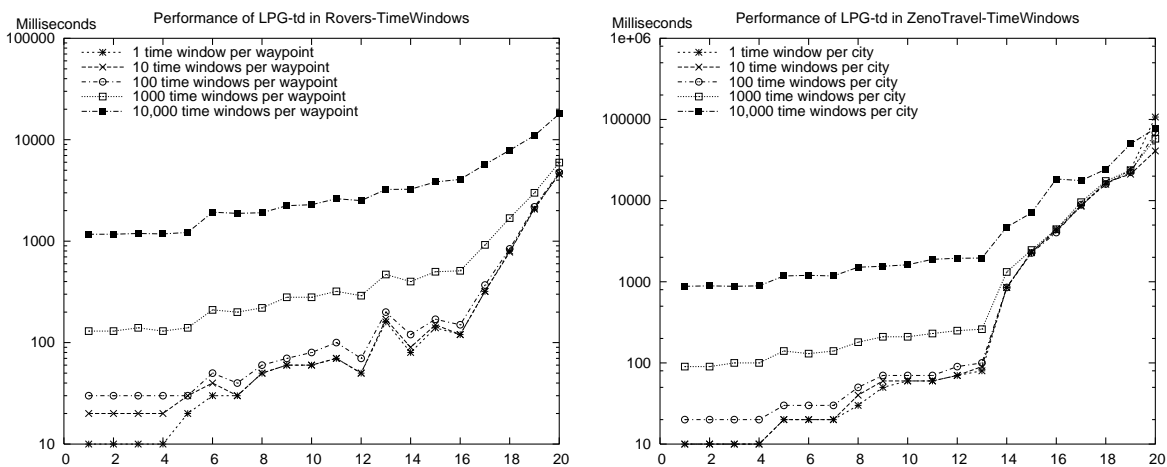

Figure 12: Performance of LPG-td in the `Rovers` and `ZenoTravel` domains extended with timed initial literals (1–10,000 time windows for each timed literal). The test problems were generated using method II. On the $x$-axis we have the problem names simplified by numbers; on the $y$-axis we have CPU-milliseconds (logarithmic scale).

The table also gives information about the average number of DTPs (i.e., search steps) generated during planning, indicating how many of them are satisfiable (indicated with "Sat. DTPs").

Table 3 gives the CPU-time required by our temporal reasoning techniques implemented in LPG-td (Solve-DTP$^+$) and by TSAT$_{++}$ (Armando, Castellini, Giunchiglia, & Maratea, 2004), a state-of-the-art general DTP solver. The DTPs considered here are the same as those of Table 2, i.e., the sets of the temporal constraints in the TDA-graph at each search step of the planning process. It should be noted that the comparison of Solve-DTP$^+$ and TSAT++ is *by no means* intended to determine which one is better than the other. Indeed TSAT++ was developed to manage a much larger class of DTPs. However, to the best of our knowledge there exists no other more specialized DTP-solver handling scheduling constraints that we could have used. The goal of this comparison is to experimentally show that existing general DTP solvers, although designed to work efficiently in the general case, are not adequate for managing the class of DTPs that arise in our planning framework. Hence, it is important to develop more specialized techniques which, as empirically demonstrated by the results of Table 3, can be much more efficient. For instance, consider problem 15 in the `Rovers` domain. As indicated by the last column of Table 2, LPG-td solves this problem with 77 search steps, which defines 77 DTPs. The data in Table 3 show that the total CPU-time spent by LPG-td for solving all these temporal reasoning problems is negligible ($< 10^{-6}$ seconds), while TSAT++ requires 16.8 CPU-seconds in total (note that the whole temporal *planning* problem is solved by LPG-td in only 0.25 seconds).[15] Overall, our specialized temporal reasoning technique is several orders of magnitude faster than an efficient general DTP, in terms of both CPU-time for solving a single DTP, and CPU-time for solving all the DTPs that are generated during planning.

---

14. The classical DTP-translation of a scheduling constraint contains an exponential number of disjuncts with respect to the number of time windows in the scheduling constraint. For example, let $q$ be a timed precondition of $a$ and $W_q = \{[25, 50), [75, 125)\}$. The scheduling constraint of $a$ determined by $q$ is translated into four classical DTP constraints ($a_s$ abbreviates $a_{start}$): $(a_s^+ - a^- \leq -25 \vee a_s^+ - a^- \leq -75)$, $(a_s^+ - a^- \leq -25 \vee a^+ - a_s^- \leq 125)$, $(a^+ - a_s^- \leq 50 \vee a_s^+ - a^- \leq -75)$, $(a^+ - a_s^- \leq 50 \vee a^+ - a_s^- \leq 125)$.

| Problems | Variables | | SC with 10 windows (DC) | | DTPs |
|---|---|---|---|---|---|
| | max | mean | max | mean | (Sat. DTPs) |
| **Rovers** | | | | | |
| Problem 1 | 28 | 18.4 | 1 (1024) | 0.13 (136.5) | 15 (15) |
| Problem 5 | 56 | 30.0 | 2 (2048) | 0.33 (341.3) | 27 (27) |
| Problem 10 | 94 | 65.8 | 2 (2048) | 1.41 (1447) | 104 (47) |
| Problem 15 | 98 | 58.8 | 3 (3072) | 1.01 (1037) | 77 (55) |
| Problem 20 | 206 | 105.0 | 4 (4096) | 1.45 (1489) | 108 (108) |
| **ZenoTravel** | | | | | |
| Problem 1 | 8 | 6 | 1 (1024) | 0.33 (341.3) | 3 (3) |
| Problem 5 | 36 | 20 | 3 (3072) | 0.88 (910.2) | 18 (18) |
| Problem 10 | 114 | 83.4 | 16 (16,384) | 10.5 (10,769) | 1162 (175) |
| Problem 15 | 172 | 122.4 | 24 (24,576) | 16.3 (16,673) | 291 (128) |
| Problem 20 | 282 | 194.6 | 42 (43,008) | 24.9 (25,536) | 750 (637) |

Table 2: Characteristics of the DTPs generated during planning by LPG-td when solving some problems in the `Rovers` and `ZenoTravel` domains: maximum/mean number of variables (2nd/3rd columns); maximum/mean number of scheduling constraints ("SC") and of non-unary disjunctions ("DC") in their DTP-form translation (4th/5th columns); number of DTPs and of satisfiable DTPs solved by LPG-td (6th column).

---

15. The CPU-time of TSAT++ includes neither the generation of the explicit (classical) DTPs from the TDA-graph, nor the parsing time. Moreover, while TSAT++ only decides satisfiability of the input DTPs, Solve-DTP$^+$ also finds a schedule that is optimal, if the DTP is satisfiable.

| Problems | CPU-seconds for Temporal Reasoning | | | | | | Total CPU-Time of LPG-td |
|---|---|---|---|---|---|---|---|
| | Solve-DTP$^+$ | | | TSAT$_{++}$ | | | |
| | max | mean | total | max | mean | total | |
| **Rovers** | | | | | | | |
| Problem 1 | $< 10^{-6}$ | $< 10^{-6}$ | $< 10^{-6}$ | 0.005 | 0.002 | 0.09 | 0.02 |
| Problem 5 | $< 10^{-6}$ | $< 10^{-6}$ | $< 10^{-6}$ | 0.045 | 0.002 | 0.14 | 0.03 |
| Problem 10 | $< 10^{-6}$ | $< 10^{-6}$ | $< 10^{-6}$ | 0.54 | 0.039 | 12.7 | 0.30 |
| Problem 15 | $< 10^{-6}$ | $< 10^{-6}$ | $< 10^{-6}$ | 0.54 | 0.028 | 16.8 | 0.25 |
| Problem 20 | 0.01 | 0.0008 | 0.03 | 3.17 | 0.10 | 107.1 | 3.03 |
| **ZenoTravel** | | | | | | | |
| Problem 1 | $< 10^{-6}$ | $< 10^{-6}$ | $< 10^{-6}$ | 0.001 | 0.0003 | 0.01 | 0.02 |
| Problem 5 | $< 10^{-6}$ | $< 10^{-6}$ | $< 10^{-6}$ | 0.04 | 0.004 | 0.21 | 0.05 |
| Problem 10 | 0.01 | 0.00017 | 0.2 | 2.7 | 9.8 | 6018 | 22.0 |
| Problem 15 | 0.01 | 0.00014 | 0.04 | 44.6 | 3.9 | 18,877 | 13.9 |
| Problem 20 | 0.01 | 0.00065 | 0.5 | 323.9 | 24.2 | 177,595 | 376.2 |

Table 3: Performance of Solve-DTP$^+$ and TSAT$_{++}$ for the DTPs generated during planning by LPG-td when solving some problems in the **Rovers** and **ZenoTravel** domains: maximum, mean and total CPU-seconds. The last column gives the total CPU-time of LPG-td for solving the planning problem. TSAT$_{++}$ was run using its default settings.

Finally, we experimentally tested the effectiveness of the improvements to Solve-DTP$^+$ for making the algorithm incremental that we have described at the end of Section 2 (such improvements are included in the implementation of Solve-DTP$^+$ of Table 3). In particular we observed that, for the problems of Table 3, the average CPU-time of the basic (non-incremental) version of Solve-DTP$^+$ is from one to three orders of magnitude higher than the incremental version. However, the basic version is still always significantly faster than TSAT$_{++}$ (from one to four orders of magnitude).

## 5. Related Work

Several researchers have addressed temporal reasoning in the context of the DTP framework. Some general techniques aimed at efficiently solving a DTP have been proposed (e.g., Armando et al., 2004; Tsamardinos & Pollack, 2003), but their worst-case complexity remains exponential. In Section 4, we presented some experimental results indicating that the simple use of a state-of-the-art DTP solver is not adequate for solving the subclass of DTPs that arise in our context.

Various planning approaches supporting the temporal features considered in this paper have been proposed. One of the first planners that was capable of handling predictable exogenous events is DEVISER (Vere, 1983), which was developed from NONLIN (Tate, 1977). DEVISER is a temporal partial order planner using a network of activities called a "plan network". Before starting plan generation, the plan network contains the exogenous events as explicit nodes of the network. During plan generation, the activities added to the network are ordered with respect to these scheduled events, depending on the relevance of the events for the activities. A similar explicit treatment of the exogenous events could be also adopted in the context of the action-graph representation: the initial action graph contains special action nodes representing the predicted exogenous events. However, this simple method has some disadvantages with respect to our method, that treats exogenous events at the *temporal level* of the representation rather than at the logical (causal) level. In particular, when there is a high number of timed initial literals, the explicit representation of the exogenous events in the action graph could lead to very large graphs, causing memory consumption problems and a possibly heavy CPU-time cost for the heuristic evaluation of the (possibly very large) search neighborhood.

In the late 80s and early 90s some other temporal planners handling exogenous events were developed. In general, these systems use input descriptions of the planning problem/domain that are significantly different from the PDDL descriptions accepted by modern fully-automated planners. One of the most successful among them is HSTS (Frederking & Muscettola, 1992; Muscettola, 1994), a representation and problem solving framework that provides an integrated view of planning and scheduling. HSTS represents predictable exogenous events through "non-controllable state variables". Both LPG-td and HSTS manage temporal constraints, but the two systems use considerably different approaches to temporal planning (LPG-td adopts the classical "state-transition view" of change, while HSTS adopts the "histories view" of change, Ghallab, Nau, & Traverso, 2003), and they are based on different plan representations and search techniques.

ZENO (Penberthy, 1993; Penberthy & Weld, 1994) is one of the first domain-independent planners which supports a rich class of metric-temporal features, including exogenous events. ZENO is a powerful extension of the causal-link partial-order planner UCPOP (Penberthy & Weld, 1992). However, in terms of computational performance, this planner is not competitive with more recent temporal planners.

IxTeT (Ghallab & Laruelle, 1994; Laborie & Ghallab, 1995) is another causal-link planner which uses some techniques and ideas from scheduling, temporal constraint reasoning, and graph algorithms. IxTeT supports a very expressive language for the temporal description of the actions, including timed preconditions and some features that cannot be expressed in PDDL2.2. The expressive power of the language is obtained at the cost of increased semantic complexity (Fox & Long, 2005). As observed by Ghallab, Nau and Traverso (2003), IxTeT embodies a compromise between the expressiveness of complex temporal domains, and the planning efficiency; however, this planner still remains noncompetitive with the more recent temporal planners.

Smith and Weld (1999) studied an extension of the Graphplan-style of planning for managing temporal domains. They proposed an extension of their TGP planner that makes it possible to represent predictable exogenous events. TGP supports only a subclass of the durative actions expressible in PDDL2.1, which prevents some cases of concurrency that in PDDL2.1 are admitted. TGP is an optimal planner (under the assumed conservative model of action concurrency), while LPG-td is a near-optimal (satisficing) planner. A main drawback of TGP is that it does not scale up adequately.

More recently, Edelkamp (2004) proposed a method for planning with timed initial literals that is based on compiling the action timed preconditions into a time window as-sociated with the action, defining the interval during which the action can be scheduled. He gives an efficient, polynomial algorithm based on critical path analysis for computing an optimal action schedule from sequential plans generated using the compiled representation. The techniques presented by Edelkamp assume a unique time window for each timed precondition. The techniques that we propose are more general, in the sense that our action representation treats multiple time windows associated with a timed precondition, and our temporal reasoning method computes optimal schedules for partially ordered plans preserving polynomiality.

Cresswell and Coddington (2004) proposed an extension of the LPGP planner (Long & Fox, 2003b) to handle timed initial literals, which are represented by special "deadline actions". A literal that is asserted to hold at time $t$ is represented by a deadline action starting at the time of the initial state, and having duration $t$. The deadline actions in the plan under construction are translated into particular linear inequalities that, together with other equalities and inequalities generated from the plan representation, are managed by a general linear programming solver. LPG-td uses a different representation that does not encode timed initial literals as special actions, and in which the temporal and scheduling constraints associated with the actions in the plan are managed by an efficient algorithm derived by specializing a general DTP solver.

In order to handle problems with timed initial literals in the SAPA planner (Do & Kambhampati, 2003), Do, Kambhampati and Zimmerman (2004) proposed a forward search heuristic based on relaxed plans, which are constructed by exploiting a technique similar to the time slack analysis used in scheduling (Smith & Cheng, 1993). Given a set of candidate actions for choosing an action to add to the relaxed plan under construction, this technique computes the minimum slack between each candidate action and the actions currently in the relaxed plan. The candidate action with the *highest minimum slack* is preferred. LPG-td uses a different time slack analysis, which is exploited in a different way. Our method for selecting the actions forming the relaxed plan uses the time slacks for counting the number of scheduling constraints that would be violated when adding a candidate action: we prefer the candidate actions which cause the *lowest number of violations*. Moreover, in SAPA the slack analysis is limited to the actions of the relaxed plan, while our method also considers the actions in the *real* plan under construction.

DT-POP is a recent planner (Schwartz & Pollack, 2004) extending the POP-style of planning with an action model involving disjunctive temporal constraints. The language of DT-POP is elegant and can express a rich class of temporal features, most of which can be only indirectly (and less elegantly) expressed in PDDL2.2 (Fox et al., 2004). The treatment of the temporal constraints required to manage predictable exogenous events in DT-POP appears to be less efficient than in our planner, since DT-POP uses a general DTP solver enhanced with some efficiency techniques, while LPG-td uses a polynomial solver specialized for the subclass of DTPs that arise in our representation. DT-POP handles mutex actions ("threats") by posting explicit temporal disjunctive constraints imposing disjointness of the mutex actions, while LPG-TD implicitly decides these disjunctions at search time by choosing the level of the graph where actions are inserted, and asserting the appropriate precedence constraints. Moreover, the search procedure and heuristics in DT-POP and LPG-td are significantly different.

At IPC-4, the planners that reasoned with timed initial literals are TILSAPA (Kavuluri & U, 2004), SGPLAN (Chen et al., 2004), P-MEP (Sanchez et al., 2004) and LPG-td. For the first two planners, at the time of writing, to the best of our knowledge in the available literature there is no sufficiently detailed description to clearly understand their possible similarities and differences with LPG-td about the treatment of predictable exogenous events. Regarding P-MEP, this planner uses forward state-space search guided by a relaxed plan heuristic which, differently from the relaxed plans of LPG-td, is constructed without taking account of the temporal aspects of the relaxed plan and real plan under construction (the makespan of the constructed relaxed plans is considered only for their comparative evaluation).

## 6. Conclusions

We have presented some techniques for temporal planning in domains where certain fluents are made true or false at known times by predictable exogenous events that cannot be influenced by the actions available to the planner. Such external events are present in many realistic domains, and a planner has to take them into account to guarantee the correctness of the synthesized plans, to generate plans of good or optimal quality (makespan), and to use effective search heuristics for fast planning.

In our approach, the causal structure of the plan is represented by a graph-based representation called TDA-graph, action ordering and scheduling constraints are managed by efficient constraint-based reasoning, and the plan search is based on a stochastic local search procedure. We have proposed an algorithm for managing the temporal constraints in a TDA-graph which is a specialization of a general CSP-based method for solving DTPs. The algorithm has a polynomial worst-case complexity and, when combined with our plan representation, in practice it is very efficient. We have also presented some local search techniques for temporal planning using the new TDA-graph representation. These techniques improve the accuracy of the heuristic methods adopted in the previous version of LPG, and they extend them to consider action scheduling constraints in the evaluation of the search neighborhood, which is based on relaxed temporal plans exploiting some (dynamic) reachability information.

All our techniques are implemented in the planner LPG-td. We have experimentally investigated the performance of our planner by a statistical analysis of the IPC-4 results using Wilcoxon's test. The results of this analysis show that our planner performs very well compared to other recent temporal planners supporting predictable exogenous events, both in terms of CPU-time to find a valid plan and quality of the best plan generated. Moreover, a comparison of the plans computed by LPG-td and those generated by the optimal planners of IPC-4 shows that very often LPG-td generates plans with very good or optimal quality. Finally, additional experiments indicate that our temporal reasoning techniques manage the class of DTPs that arise in our context very efficiently.

Some directions for future work on temporal planning within our framework are: an extension of the local search heuristics and temporal reasoning techniques to explicitly handle action effects with limited persistence or delays; the treatment of predictable exogenous events affecting numerical fluents in a discrete or continuous way; the development of tech-niques supporting controllable exogenous events;[16] and the management of actions with "variable" durations (Fox & Long, 2003), i.e., actions whose durations are specified only by inequalities constraining their lower or upper bounds, and whose actual duration is decided by the planner.

Moreover, we intend to study the integration into our framework of the techniques for goal partitioning and subplan composition that have been successfully used by SGPLAN (Chen et al., 2004) in IPC-4, and the application of our approach to plan revision. The latter has already been partially explored, but only for simple STRIPS domains and using less powerful search techniques (Gerevini & Serina, 2000).

## Acknowledgments

This paper is a revised and extended version of a paper appearing in the Proceedings of the Nineteenth International Joint Conference on Artificial Intelligence (Gerevini, Saetti, & Serina, 2005a). The research was supported in part by MIUR Grant ANEMONE. The work of Ivan Serina was in part carried out at the Department of Computer and Information Sciences of the University of Strathclyde (Glasgow, UK), and was supported by Marie Curie Fellowship N HPMF-CT-2002-02149. We would like to thank the anonymous reviewers for their helpful comments, and Paolo Toninelli who extended the parser of LPG-td to handle the new language features of PDDL2.2.

## Appendix A: Reachability Information

The techniques described in the paper for computing the action evaluation function use heuristic reachability information about the minimum number of actions required to reach the preconditions of each domain action ($Num\_acts$) and a lower bound on the earliest finishing time ($Eft$) of the reachable actions (the actions whose preconditions are reachable). In the following, $S(l)$ denotes the state defined by the facts corresponding to the fact nodes supported at level $l$ of the current TDA-graph. When $l = 1$, $S(l)$ represents the initial state of the planning problem ($I$).

For each action $a$, LPG-td pre-computes $Num\_acts(a, I)$, i.e., the estimated minimum number of actions required to reach the preconditions of $a$ from $I$, and $Eft(a, I)$, i.e., the estimated earliest finishing time of $a$ (if $a$ is reachable from $I$). Similarly, for each fact $f$ that is reachable from $I$, LPG-td computes the estimated minimum number of actions required to reach $f$ from $I$ ($Num\_acts(f, I)$) and the estimated earliest time when $f$ can be made true by a plan starting from $I$ ($Et(f, I)$). For $l > 1$, $Num\_acts(a, S(l))$ and $Eft(a, S(l))$ can be computed only during search, because they depend on which action nodes are in the current TDA-graph at the levels preceding $l$. Since during search many action nodes can be added and removed, and after each of these operations $Num\_acts(a, S(l))$ and $Eft(a, S(l))$ could change (if the operation concerns a level preceding $l$), it is important that they are computed efficiently.

---

16. Consider for instance a transportation domain in which a shuttle bus is at the train station for an extra run to the airport at midnight only if booked in advance. If the shuttle booking is a domain action available to the planner, then the event "night stop of the shuttle" can be controlled by the planner.

ReachabilityInformation$(I, \mathcal{O})$

   *Input*: The initial state of the planning problem under consideration $(I)$ and all ground instances
         (actions) of the operators $(\mathcal{O})$;

   *Output*: For each action $a$, an estimate of the number of actions $(Num\_acts(a, I))$ required to reach
        the preconditions of $a$ from $I$, an estimate of the earliest finishing time of $a$ from $I$ $(Eft(a, I))$.

1.    **forall** facts $f$ **do** /* the set of all facts is precomputed by the operator instantiation phase */
2.      **if** $f \in I$ **then**
3.         $Num\_acts(f, I) \leftarrow Et(f, I) \leftarrow 0;$   $Action(f, I) \leftarrow a_{start};$
4.      **else** $Num\_acts(f, I) \leftarrow Et(f, I) \leftarrow \infty;$
5.    **forall** actions $a$ **do** $Num\_acts(a, I) \leftarrow Eft(a, I) \leftarrow Lft(a) \leftarrow \infty;$
6.    $F \leftarrow I;$   $F_{new} \leftarrow I;$   $A \leftarrow \mathcal{O};$   $A_{rev} \leftarrow \emptyset;$
7.    **while** ( $F_{new} \neq \emptyset$ or $A_{rev} \neq \emptyset$ )
8.      $F \leftarrow F \cup F_{new};$   $F_{new} \leftarrow \emptyset;$   $A \leftarrow A \cup A_{rev};$   $A_{rev} \leftarrow \emptyset;$
9.      **while** $A' = \{a \in A \mid Pre(a) \subseteq F\}$ is not empty
10.        $a \leftarrow$ an action in $A';$
11.        $t \leftarrow ComputeEFT(a, \underset{f \in Pre(a)}{MAX} Et(f, I));$
12.        **if** $t < Eft(a, I)$ **then** $Eft(a, I) \leftarrow t;$
13.        $Lft(a) \leftarrow ComputeLFT(a);$
14.        **if** $Eft(a, I) \leq Lft(a)$ **then** /* $a$ can be scheduled */
15.          $ra \leftarrow$ RequiredActions$(I, Pre(a));$
16.          **if** $Num\_acts(a, I) > ra$ **then** $Num\_acts(a, I) \leftarrow ra;$
17.          **forall** $f \in Add(a)$ **do**
18.            **if** $Et(f, I) > t$ **then**
19.              $Et(f, I) \leftarrow t;$
20.              $A_{rev} \leftarrow A_{rev} \cup \{a' \in \mathcal{O} - A \mid f \in Pre(a')\};$
21.            **if** $Num\_acts(f, I) > (ra + 1)$ **then**
22.              $Num\_acts(f, I) \leftarrow ra + 1;$   $Action(f, I) \leftarrow a;$
23.          $F_{new} \leftarrow F_{new} \cup Add(a) - F;$
24.        $A \leftarrow A - \{a\};$

RequiredActions$(I, G)$

   *Input*: A set of facts $I$ and a set of action preconditions $G$;
   *Output*: An estimate of the min number of actions required to achieve all facts in $G$ from $I$ $(ACTS)$.

1.    $ACTS \leftarrow \emptyset;$
2.    $G \leftarrow G - I;$
3.    **while** $G \neq \emptyset$
4.      $g \leftarrow$ an element of $G;$
5.      $a \leftarrow Action(g, I);$
6.      $ACTS \leftarrow ACTS \cup \{a\};$
7.      $G \leftarrow G \cup Pre(a) - I - \bigcup_{b \in ACTS} Add(b);$
8.    **return**$(|ACTS|).$

Figure 13: Algorithms for computing heuristic information about the search cost and the
         time for reaching a set of facts $G$ from $I$.

Figure 13 gives ReachabilityInformation, the algorithm used by LPG-td for computing $Num\_acts(a, I)$, $Eft(a, I)$, $Num\_acts(f, I)$ and $Et(f, I)$. ReachabilityInformation is similar to the reachability algorithm used by the version of LPG that took part in 2002 planning competition (LPG-IPC3), but with some significant differences. The main differences are:

(i) in order to estimate the earliest finishing time of the domain actions, ReachabilityInformation takes into account the scheduling constraints, which were not considered in the previous version of the algorithm;

(ii) the algorithm used by LPG-IPC3 applies each domain action at most once, while ReachabilityInformation can apply them more than once.

Notice that (i) improves the accuracy of the estimated finishing time of the actions ($Eft$), which is an important piece of information used during the search neighborhood evaluation for selecting the actions forming the temporal relaxed plans (see Section 3). Moreover, (i) allows us to identify some domain actions that cannot be scheduled during the time windows associated with their timed preconditions, and so these can be pruned away.

Regarding (ii), during the forward process of computing the reachability information, an action is *re*-applied whenever the estimated earliest time of one of its preconditions has been decreased. This is important for two reasons. On one hand, reconsidering actions already applied is useful because it can lead to a better estimate of the action finishing times; on the other hand, this is also necessary to guarantee the correctness of the reachability algorithm. The latter is because, if we overestimate the earliest finishing time of an action with a scheduling constraint, then we could incorrectly conclude that the action cannot be scheduled (and so we would consider the action inapplicable). But if this action is necessary in any valid plan, then the incorrect estimate of its earliest finishing time could lead to the incorrect conclusion that the planning problem is unsolvable. In other words, the estimated finishing time of an action with a scheduling constraint should be a *lower bound* of its actual earliest finishing time.

ReachabilityInformation could be used to update $Num\_acts(a, S(l))$ and $Eft(a, S(l))$ after each action insertion/removal, for any $l > 1$ (when $l > 1$, instead of $I$, in input the algorithm has $S(l)$). However, in order to make the updating process more efficient, the revision is done in a more selective focused way. Instead of revising the reachability information after each graph modification (search step), we do so *before* evaluating the search neighborhood and choosing the estimated best modification. Specifically, if we are repairing the flawed level $l$, we update only the reachability information for the actions and facts at the levels *preceding* $l$ that have not been updated yet. (For instance, suppose that at the $i$th search step we add an action to level 5, and that at the $(i + 1)$th step we add another action at level 10. At the $(i + 1)$th step we need to consider updating only the reachability information at levels 6–10, since this information at levels 1–5 has already been updated by the $i$th step.) This is sufficient because the search neighborhood for repairing the flawed level under consideration ($l$) can contain only the graph modifications concerning the levels preceding $l$.

Before describing the steps of ReachabilityInformation, we need to introduce some notation. $Add(a)$ denotes the set of the positive effects of $a$; $Pre(a)$ denotes the set of the (non-timed) preconditions of $a$; $A_{rev}$ denotes the set of the actions already applied whose reachability could be revised because the estimated earliest time of some of their preconditions has been revised after their application. Given an action node $a$ and its "current" earliest start time $t$ computed as the maximum over the earliest times at which its preconditions are reachable, $ComputeEFT(a, t)$ is a function computing the *earliest finishing time* $\tau$ of $a$ that is consistent with the scheduling constraint of $a$ (if any) and such that $t + Dur(a) \leq \tau$.[17] $ComputeLFT(a)$ is a function computing the *latest finishing time* of the action $a$, i.e., it returns the upper bound of the last time window during which $a$ can be scheduled (if one exists), while it returns $\infty$ if $a$ has no timed precondition.

For example, let $a$ be an action such that all its preconditions are true in the initial state $I$ (i.e., $t = 0$), the duration of $a$ is 50, and $a$ has a scheduling constraint imposing that the action is executed during the interval $[25, 100]$. $ComputeEFT(a, t)$ returns 75, while $ComputeLFT(a, t)$ returns 100. Thus, the scheduling constraint of $a$ can be satisfied. On the contrary, if the earliest start time of $a$ is 500, then $ComputeEFT(a, t)$ returns 550 and $a$ cannot be scheduled during $[25, 100]$.

For the sake of clarity, first we describe the steps of ReachabilityInformation used to derive $Num\_acts$, and then we comment on those for the computation of $Eft$. In steps 1–4, for every fact $f$, the algorithm initializes $Num\_acts(f, I)$ to 0, if $f \in I$, and to $\infty$ otherwise (indicating that $f$ is not reachable); while, in step 5, $Num\_acts(a, I)$ is initialized to $\infty$ (indicating that $a$ is not reachable from $I$). Then, in steps 7–24 the algorithm iteratively constructs the set $F$ of the facts that are reachable from $I$, starting with $F = I$, and terminating when $F$ cannot be further extended and the set $A_{rev}$ of the actions to reconsider is empty. The set $A$ of the available actions is initialized to the set of all possible actions (step 6); $A$ is reduced by $a$ after its application (step 24), and it is augmented by the set of actions $A_{rev}$ (step 8) after each action application. When we modify the estimated time at which a precondition of an action $a$ becomes reachable, $a$ is added to $A_{rev}$ (step 20). The internal while-loop (steps 9–24) applies the actions in $A$ to the current $F$, possibly deriving a new set of facts $F_{new}$ in step 23. If $F_{new}$ or $A_{rev}$ are not empty, then $F$ is extended with $F_{new}$, $A$ is extended with $A_{rev}$, and the internal loop is repeated. When an action $a$ in $A'$ (the subset of actions currently in $A$ that are applicable to $F$) is applied, the reachability information for its effects are revised as follows. First we estimate the minimum number $ra$ of actions required to achieve $Pre(a)$ from $I$ using the subroutine RequiredActions (step 15). Then we use $ra$ to possibly update $Num\_acts(a, I)$ and $Num\_acts(f, I)$ for any effect $f$ of $a$ (steps 15–16, 21–22). If the number of actions required to achieve the preconditions of $a$ is lower than the current value of $Num\_acts(a, I)$, then $Num\_acts(a, I)$ is set to $ra$. Moreover, if the application of $a$ leads to a lower estimate of $f$, i.e., if $ra + 1$ is less than the current value of $Num\_acts(f, I)$, then $Num\_acts(f, I)$ is set to $ra + 1$. In addition, a data structure indicating the current "best" action to achieve $f$ from $I$ ($Action(f, I)$) is set to $a$ (step 22). This information is used by the subroutine RequiredActions.

For any fact $f$ in the initial state, the value of $Action(f, I)$ is $a_{start}$ (step 3). The subroutine RequiredActions is the same as the one in the reachability algorithm of LPG-IPC3. The subroutine uses $Action$ to derive $ra$ through a backward process starting from the input set of action preconditions ($G$), and ending when $G \subseteq I$. The subroutine incrementally constructs a set of actions ($ACTS$) achieving the facts in $G$ and the preconditions of the actions already selected (using $Action$). At each iteration the set $G$ is revised by adding the preconditions of the last action selected, and removing the facts belonging to $I$ or to the effects of actions already selected (step 7). Termination of RequiredActions is guaranteed because every element of $G$ is reachable from $I$.

---

17. If there is no scheduling constraint associated with $a$, or the existing scheduling constraints cannot be satisfied by starting the action at $t$, then $ComputeEFT(a, t)$ returns $t + Dur(a)$.

We now briefly describe the computation of the temporal information. $Eft(a, I)$, is computed in a way similar to $Num\_acts(a, I)$. In steps 1–4, ReachabilityInformation initializes the estimated earliest time $(Et(f, I))$ when a fact $f$ becomes reachable to 0, if $f \in I$, and to $\infty$ otherwise; moreover, the algorithm sets $Eft(a, I)$ and $Lft(a, I)$ to $\infty$. Then, at every application of an action $a$ in the forward process described above, we estimate the earliest finishing time $Eft$ by adding the duration of $a$ to the (current) maximum estimated earliest time of the preconditions of $a$, and by taking into account the scheduling constraints of $a$ using $ComputeEFT(a)$ (step 11). In addition, we compute the latest finishing time $Lft$ of $a$ using $ComputeLFT(a)$ (step 13). When the earliest finishing time of an action $a$ is greater than its latest finishing time, the timed preconditions of $a$ cannot be satisfied from $I$, and so steps 15–23 are not executed (see the if-statement of step 14). For any effect $f$ of $a$ with a current temporal value higher than the earliest finishing time $t$ of $a$, steps 18–19 set $Et(f, I)$ to $t$, and step 20 adds $a$ in $A_{rev}$ (because we have decreased the estimated earliestx time of $f$, and this revision could decrease the estimated start time of an action with precondition $f$).

## Appendix B: Wilcoxon Test for the Metric-Temporal Domains of IPC-4

In this appendix, we present the results of the Wilcoxon sign-rank test on the performance of LPG-td and the other satisficing IPC-4 planners that attempted the metric-temporal domains. The performance is evaluated both in terms of CPU-times and plan quality.

Each cell in the first two tables gives the result of a comparison between the performance of LPG-td and another IPC-4 planner. When the number of samples is sufficiently large, the T-distribution used by the Wilcoxon test is approximatively a normal distribution. Hence, in each cell of the Figure we give the $z$-value and the $p$-value characterizing the normal distribution. The higher the $z$-value, the more significant the difference of the performance is. The $p$-value represents the level of significance in the difference of the performance. We use a confidence level of 99.9%; therefore, if the $p$-value is lower than 0.001, then the performance of the two planners is statistically different. When this information appears on the left (right) side of the cell, the first (second) planner named in the title of the cell performs better than the other. For the analysis comparing the CPU-time, the value under each cell is the number of the problems solved by at least one planner; while for the analysis comparing the plan quality, it is the number of problems solved by both the planners.

The pictures under the tables show the partial order of the performance of the compared planners in terms of CPU-time and plan quality. A solid edge from a planner A to another planner B (or a cluster of planners B) indicates that the performance of A is statistically different from the performance of B, and that A performs better than B (every planner in B). A dashed edge from A to B indicates that A is better than B a significant number of times, but there is not significant Wilcoxon relationship between them at a confidence level of 99.9%.

| Analysis of CPU-Time | | | | | | | |
|---|---|---|---|---|---|---|---|
| LPG-td.s *vs* CRIKEY | | LPG-td.s *vs* P-MEP | | LPG-td.s *vs* SGPLAN | | LPG-td.s *vs* TILSAPA | |
| 11.275 | | 11.132 | | | 0.387 | 12.324 | |
| < 0.001 | | < 0.001 | | | (0.699) | < 0.001 | |
| 169 | | 215 | | 513 | | 136 | |

| Analysis of Plan Quality | | | | | | | |
|---|---|---|---|---|---|---|---|
| LPG-td.bq *vs* CRIKEY | | LPG-td.bq *vs* P-MEP | | LPG-td.bq *vs* SGPLAN | | LPG-td.bq *vs* TILSAPA | |
| 10.500 | | 4.016 | | 16.879 | | 6.901 | |
| < 0.001 | | < 0.001 | | < 0.001 | | < 0.001 | |
| 173 | | 21 | | 452 | | 63 | |

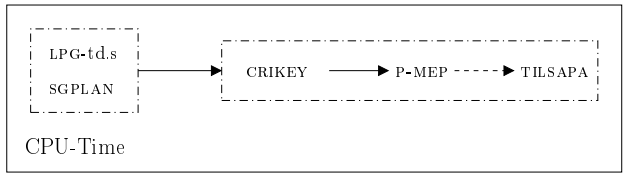

CPU-Time

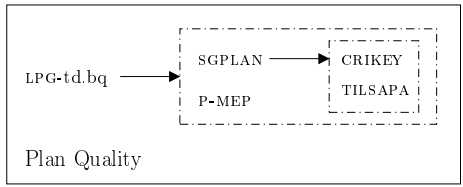

Plan Quality

A ———▶ B :  A is consistently better than B

A ---▶ B :  A is better than B a significant number of times
            (confidence level 94.78%)